\documentclass[journal]{IEEEtran}
\ifCLASSINFOpdf
\else
\fi

\usepackage{lineno}
\usepackage{amsmath,amssymb,amsfonts}
\usepackage{enumitem}
\usepackage{algorithm}
\usepackage{algorithmic}
\usepackage{textcomp}
\usepackage{multirow}
\usepackage{multicol}
\usepackage{longtable}
\usepackage{lipsum}
\usepackage{float}
\usepackage{dblfloatfix}
\usepackage{mathtools}
\usepackage{xcolor}
\usepackage{array}
\usepackage{graphicx}

\ifCLASSOPTIONcompsoc
\usepackage[caption=false,font=normalsize,labelfon
t=sf,textfont=sf]{subfig}
\else
\usepackage[caption=false,font=footnotesize]{subfig}
\fi

\newcommand{\PreserveBackslash}[1]{\let\temp=\\#1\let\\=\temp}
\newcolumntype{C}[1]{>{\PreserveBackslash\centering}p{#1}}
\newcolumntype{R}[1]{>{\PreserveBackslash\raggedleft}p{#1}}
\newcolumntype{L}[1]{>{\PreserveBackslash\raggedright}p{#1}}

\usepackage{scalerel}
\usepackage{tikz}
\usetikzlibrary{svg.path}

\definecolor{orcidlogocol}{HTML}{A6CE39}
\tikzset{
	orcidlogo/.pic={
		\fill[orcidlogocol] svg{M256,128c0,70.7-57.3,128-128,128C57.3,256,0,198.7,0,128C0,57.3,57.3,0,128,0C198.7,0,256,57.3,256,128z};
		\fill[white] svg{M86.3,186.2H70.9V79.1h15.4v48.4V186.2z}
		svg{M108.9,79.1h41.6c39.6,0,57,28.3,57,53.6c0,27.5-21.5,53.6-56.8,53.6h-41.8V79.1z M124.3,172.4h24.5c34.9,0,42.9-26.5,42.9-39.7c0-21.5-13.7-39.7-43.7-39.7h-23.7V172.4z}
		svg{M88.7,56.8c0,5.5-4.5,10.1-10.1,10.1c-5.6,0-10.1-4.6-10.1-10.1c0-5.6,4.5-10.1,10.1-10.1C84.2,46.7,88.7,51.3,88.7,56.8z};
	}
}

\newcommand\orcidicon[1]{\href{https://orcid.org/#1}{\mbox{\scalerel*{
				\begin{tikzpicture}[yscale=-1,transform shape]
				\pic{orcidlogo};
				\end{tikzpicture}
			}{|}}}}
		
\usepackage[hidelinks]{hyperref}

\begin{document}
%
\title{An Effective Multi-Resolution Hierarchical Granular Representation based Classifier using General Fuzzy Min-Max Neural Network}
%
%
%

\author{Thanh Tung Khuat$^{\textsuperscript{\orcidicon{0000-0002-6456-8530}}}$,~Fang Chen$^{\textsuperscript{\orcidicon{0000-0003-4971-8729}}}$,~and~Bogdan Gabrys$^{\textsuperscript{\orcidicon{0000-0002-0790-2846}}}$,~\IEEEmembership{Senior Member,~IEEE}

\thanks{T.T. Khuat (email: thanhtung.khuat@student.uts.edu.au) and B. Gabrys (email: Bogdan.Gabrys@uts.edu.au) are with Advanced Analytics Institute, Faculty of Engineering and Information Technology, University of Technology Sydney, Ultimo, NSW 2007, Australia.
	
F. Chen (email: Fang.Chen@uts.edu.au) is with Data Science Centre, Faculty of Engineering and Information Technology, University of Technology Sydney, Ultimo, NSW 2007, Australia.}
}

%
%

\markboth{}
{Khuat \MakeLowercase{\textit{et al.}}: An Effective Multi-Resolution Hierarchical Granular Representation based Classifier using GFMM}

%



\maketitle

\begin{abstract}
Motivated by the practical demands for simplification of data towards being consistent with human thinking and problem solving as well as tolerance of uncertainty, information granules are becoming important entities in data processing at different levels of data abstraction. This paper proposes a method to construct classifiers from multi-resolution hierarchical granular representations (MRHGRC) using hyperbox fuzzy sets. The proposed approach forms a series of granular inferences hierarchically through many levels of abstraction. An attractive characteristic of our classifier is that it can maintain a high accuracy in comparison to other fuzzy min-max models at a low degree of granularity based on reusing the knowledge learned from lower levels of abstraction. In addition, our approach can reduce the data size significantly as well as handle the uncertainty and incompleteness associated with data in real-world applications. The construction process of the classifier consists of two phases. The first phase is to formulate the model at the greatest level of granularity, while the later stage aims to reduce the complexity of the constructed model and deduce it from data at higher abstraction levels. Experimental analyses conducted comprehensively on both synthetic and real datasets indicated the efficiency of our method in terms of training time and predictive performance in comparison to other types of fuzzy min-max neural networks and common machine learning algorithms.
\end{abstract}


%
\IEEEpeerreviewmaketitle

%
%
%
%

\section{Introduction}
\label{intro}
\IEEEPARstart{H}{ierarchical} problem solving, where the problems are analyzed in a variety of granularity degrees, is a typical characteristic of the human brain  \cite{Wang17}. Inspired by this ability, granular computing was introduced. One of the critical features of granular computing is to model the data as high-level abstract structures and to tackle problems based on these representations similar to structured human thinking \cite{Morente-Molinera17}. Information granules (IGs) \cite{Zadeh97} are underlying constructs of the granular computing. They are abstract entities describing important properties of numeric data and formulating knowledge pieces from data at a higher abstraction level. They play a critical role in the concise description and abstraction of numeric data \cite{Pedrycz15}. Information granules have also contributed to quantifying the limited numeric precision in data \cite{Pedrycz14}.

Utilizing information granules is one of the problem-solving methods based on decomposing a big problem into sub-tasks which can be solved individually. In the world of big data, one regularly departs from specific data entities and discover general rules from data via encapsulation and abstraction. The use of information granules is meaningful when tackling the five Vs of big data \cite{Xu18}, i.e., volume, variety, velocity, veracity, and value. Granulation process gathering similar data together contributes to reducing the data size, and so the volume issue is addressed. The information from many heterogeneous sources can be granulated into various granular constructs, and then several measures and rules for uniform representation are proposed to fuse base information granules as shown in \cite{Xuweihua17}. Hence, the data variety is addressed. Several studies constructed the evolving information granules to adapt to the changes in the streams of data as in \cite{Al-Hmouz18a}. The variations of information granules in a high-speed data stream assist in tackling the velocity problem of big data. The process of forming information granules is often associated with the removal of outliers and dealing with incomplete data \cite{Xu18}; thus the veracity of data is guaranteed. Finally, the multi-resolution hierarchical architecture of various granular levels can disregard some irrelevant features but highlight facets of interest \cite{Chen14}. In this way, the granular representation may help with cognitive demands and capabilities of different users.

A multi-dimensional hyperbox fuzzy set is a fundamental conceptual vehicle to represent information granules. Each fuzzy min-max hyperbox is determined by the minimum and maximum points and a fuzzy membership function. A classifier can be built from the hyperbox fuzzy sets along with an appropriate training algorithm. We can extract a rule set directly from hyperbox fuzzy sets or by using it in combination with other methods such as decision trees \cite{Khuat19} to account for the predictive results. However, a limitation of hyperbox-based classifiers is that their accuracy at the low level of granularity (corresponding to large-sized hyperboxes) decreases. In contrast, classifiers at the high granularity level are more accurate, but the building process of classifiers at this level is time-consuming, and it is difficult to extract the rule set interpretable for predictive outcomes because of the high complexity of resulting models. Hence, it is desired to construct a simple classifier with high accuracy. In addition, we expect to observe the change in the predictive results at different data abstraction levels. This paper proposes a method of constructing a high-precision classifier at the high data abstraction level based on the knowledge learned from lower abstraction levels. On the basis of classification errors on the validation set, we can predict the change in the accuracy of the constructed classifier on unseen data, and we can select an abstraction level satisfying both acceptable accuracy and simple architecture on the resulting classifier. Furthermore, our method is likely to expand for large-sized datasets due to the capability of parallel execution during the constructing process of core hyperboxes at the highest level of granularity. In our method, the algorithm starts with a relatively small value of maximum hyperbox size ($\theta$) to produce base hyperbox fuzzy sets, and then this threshold is increased in succeeding levels of abstraction whose inputs are the hyperbox fuzzy sets formed in the previous step. By using many hierarchical resolutions of granularity, the information captured in earlier steps is transferred to the classifier at the next level. Therefore, the classification accuracy is still maintained at an acceptable value when the resolution of training data is low.

Data generated from complex real-world applications frequently change over time, so the machine learning models used to predict behaviors of such systems need the efficient online learning capability. Many studies considered the online learning capability when building machine learning models such as \cite{Simpson92, Gabrys00, Rubio09, Zhang17state, Rubio17, Cheng2018}, and \cite{Rubio19}. Fuzzy min-max neural networks proposed by Simpson \cite{Simpson92} and many of its improved variants only work on the input data in the form of points. In practice, due to the uncertainty and some abnormal behaviors in the systems, the input data include not only crisp points but also intervals. To address this problem, Gabrys and Bargiela \cite{Gabrys00} introduced a general fuzzy min-max (GFMM) neural network, which can handle both fuzzy and crisp input samples. By using hyperbox fuzzy sets for the input layer, this model can accept the input patterns in the granular form and process data at a high-level abstract structure. As a result, our proposed method used a similar mechanism as in the general fuzzy min-max neural network to build a series of classifiers through different resolutions, where the small-sized resulting hyperbox fuzzy sets generated in the previous step become the input to be handled at a higher level of abstraction (corresponding to a higher value of the allowable hyperbox size). Going through different resolution degrees, the valuable information in the input data is fuzzified and reduced in size, but our method helps to preserve the amount of knowledge contained in the original datasets. This capability is illustrated via the slow decline in the classification accuracy. In some cases, the predictive accuracy increases at higher levels of abstraction because the noise existing in the detailed levels is eliminated.

Building on the principles of developing GFMM classifiers with good generalization performance discussed in \cite{Gabrys04}, this paper employs different hierarchical representations of granular data with various hyperbox sizes to select a compact classifier with acceptable accuracy at a high level of abstraction. Hierarchical granular representations using consecutive maximum hyperbox sizes form a set of multi-resolution hyperbox-based models, which can be used to balance the trade-off between efficiency and simplicity of the classifiers. A model with high resolution corresponds to the use of a small value of maximum hyperbox size, and vice versa. A choice of suitable resolution level results in better predictive accuracy of the generated model. Our main contributions in this paper can be summarized as follows:

\begin{itemize}
	\item We propose a new data classification model based on the multi-resolution of granular data representations in combination with the online learning ability of the general fuzzy min-max neural network.
	\item The proposed method is capable of reusing the learned knowledge from the highest granularity level to construct new classifiers at higher abstraction levels with the low trade-off between the simplification and accuracy.
	\item The efficiency and running time of the general fuzzy min-max classifier are significantly enhanced in the proposed algorithm.
	\item Our classifier can perform on large-sized datasets because of the parallel execution ability.
	\item Comprehensive experiments are conducted on synthetic and real datasets to prove the effectiveness of the proposed method compared to other approaches and baselines.
\end{itemize}

The rest of this paper is organized as follows. Section \ref{prelimi} presents existing studies related to information granules as well as briefly describes the online learning version of the general fuzzy min-max neural network. Section \ref{method} shows our proposed method to construct a classifier based on data granulation. Experimental configuration and results are presented in Section \ref{experiment}. Section \ref{conclu} concludes the main findings and discusses some open directions for future works.

\section{Preliminaries}\label{prelimi}
\subsection{Related Work}
There are many approaches to representing information granules \cite{Pedrycz18}. Several typical methods include intervals \cite{Moore09}, fuzzy sets \cite{Zadeh65}, shadowed sets \cite{Pedrycz05}, and rough sets \cite{Pawlak07}. Our study only focuses on fuzzy sets and intervals. Therefore, related works only mention the granular representation using these two methods.

The existing studies on the granular data representation have deployed a specific clustering technique to find representative prototypes, and then build information granules from these prototypes and optimize the constructed granular descriptors. The principle of justifiable granularity \cite{Pedrycz13} has been usually utilized to optimize the construction of information granules from available experimental evidence. This principle aims to make a good balance between coverage and specificity properties of the resulting granule concerning available data. The coverage property relates to how much data is located inside the constructed information granule, whereas the specificity of a granule is quantified by its length of the interval such that the shorter the interval, the better the specificity. Pedrycz and Homenda \cite{Pedrycz13} made a compromise between these two characteristics by finding the parameters to maximize the product of the coverage and specificity.

Instead of just stopping at the numeric prototypes, partition matrices, or dendrograms for data clustering, Pedrycz and Bargiela \cite{Pedrycz12} offered a concept of granular prototypes to capture more details of the structure of data to be clustered. Granular prototypes were formed around the resulting numeric prototypes of clustering algorithms by using some degree of granularity. Information granularity is allocated to each numeric prototype to maximize the quality of the granulation-degranulation process of generated granules. This process was also built as an optimization problem steered by the coverage criteria, i.e., maximization of the original number of data included in the information granules after degranulation. 

In \cite{Pedrycz14}, Pedrycz developed an idea of granular models derived from the establishment of optimal allocation of information granules. The authors gave motivation and plausible explanations in bringing the numeric models to the next abstraction levels to form granular models. In the realization flow of the general principles, Pedrycz et~al. \cite{Pedrycz15} introduced a holistic process of constructing information granules through a two-phase procedure in a general framework. The first phase focuses on formulating numeric prototypes using fuzzy C-means, and the second phase refines each resulting prototype to form a corresponding information granule by employing the principle of justifiable granularity.

When the problem becomes complicated, one regularly splits it into smaller sub-tasks and deal with each sub-task on a single level of granularity. These actions give rise to the appearance of multi-granularity computing, which aims to tackle problems from many levels of different IGs rather than just one optimal granular layer. Wang et~al. \cite{Wang17} conducted a review on previous studies of granular computing and claimed that multi-granularity joint problem resolving is a valuable research direction to enhance the quality and efficiency of solutions based on using multiple levels of information granules rather than only one granular level. This is the primary motivation for our study to build suitable classifiers from many resolutions of granular data representations.

All the above methods of building information granules are based on the clustering techniques and affected by a pre-determined parameter, i.e., the number of clusters. The resulting information granules are only summarization of the original data at a higher abstraction level, and they did not use the class information in the constructing process of granules. The authors have not used the resulting granules to deal with classification problems either. Our work is different from these studies because we propose a method to build classifiers from various abstraction levels of data using the hyperbox fuzzy sets while maintaining the reasonable stability of classification results. In addition, our method can learn useful information from data through an online approach and the continuous adjustment of the existing structure of the model.

In the case of formulating models in a non-stationary environment, it is essential to endow them with some mechanisms to deal with the dynamic environment. In \cite{Sahel07}, Sahel et al. assessed two adaptive methods to tackle data drifting problems, i.e., retraining models using evolving data and deploying incremental learning algorithms. Although these approaches improved the accuracy of classifiers compared to non-adaptive learners, the authors indicated a great demand on building robust techniques with high reliability for dynamic operating environments. To meet the continuous changing in data and the adaptation of the analytic system to this phenomenon, Peters and Weber \cite{Peters16} suggested a framework, namely Dynamic Clustering Cube, to classify dynamic granular clustering methods. Al-Hmouz et~al. \cite{Al-Hmouz18a} introduced evolvable data models through the dynamic changing of temporal or spatial IGs. The number of information granules formed from prototypes of data can increase or decrease through merging or splitting existing granules according to the varying complexity of data streams. In addtion to the ability to merge existing hyperboxes for the construction of granular hierarchies of hyperboxes, our proposed method also has the online learning ability, so it can be used to tackle classification problems in a dynamic environment.

Although this study is built based on the principles of GFMM classifiers, it differs from the GFMM neural network with adaptive hyperbox sizes \cite{Gabrys00}. In \cite{Gabrys00}, the classifier was formed at the high abstraction level with large-sized hyperboxes and then repeating many times the process of traversing entire training dataset to build additional hyperboxes at lower abstraction levels with the aim of covering patterns missed by large-sized hyperboxes due to the contraction procedure. This operation can make the final classifier complex with a large number of hyperboxes at different levels of granularity coexisting in a single classifier, and overfitting phenomenon on the training set is more likely to happen. In contrast, our method begins with the construction process of the classifier at the highest resolution of training patterns with small-sized hyperboxes to capture detailed information and relationships among data points located in the vicinity of each other. After that, at higher levels of abstraction, we do not use the data points from the training set. Instead, we reuse the hyperboxes generated from the preceding step. For each input hyperbox, the membership value with the hyperboxes in the current step are computed, and if the highest membership degree is larger than a pre-defined threshold, the aggregation process is performed to form hyperboxes with higher abstraction degrees. Our research is also different from the approach presented in \cite{Gabrys02agglo}, where the incremental algorithm was employed to reduce the data complexity by creating small-sized hyperboxes, and then these hyperboxes were used as training inputs of an agglomerative learning algorithm with a higher abstraction level. The method in \cite{Gabrys02agglo} only constructs the classifier with two abstraction levels, while our algorithm can generate a series of classifiers at hierarchical resolutions of abstraction levels. In addition, the agglomerative learning in \cite{Gabrys02agglo} is time-consuming, especially in large-sized datasets. Therefore, when the number of generated hyperboxes using the incremental learning algorithm on large-sized training sets is large, the agglomerative learning algorithm takes a long time to formulate hyperboxes. On the contrary, our method takes advantage of the incremental learning ability to build rapidly classifiers through different levels of the hierarchical resolutions.

\subsection{General Fuzzy Min-Max Neural Network} \label{gfmm}
\begin{figure}[!ht]
	\centering
	\includegraphics[width=0.65\linewidth]{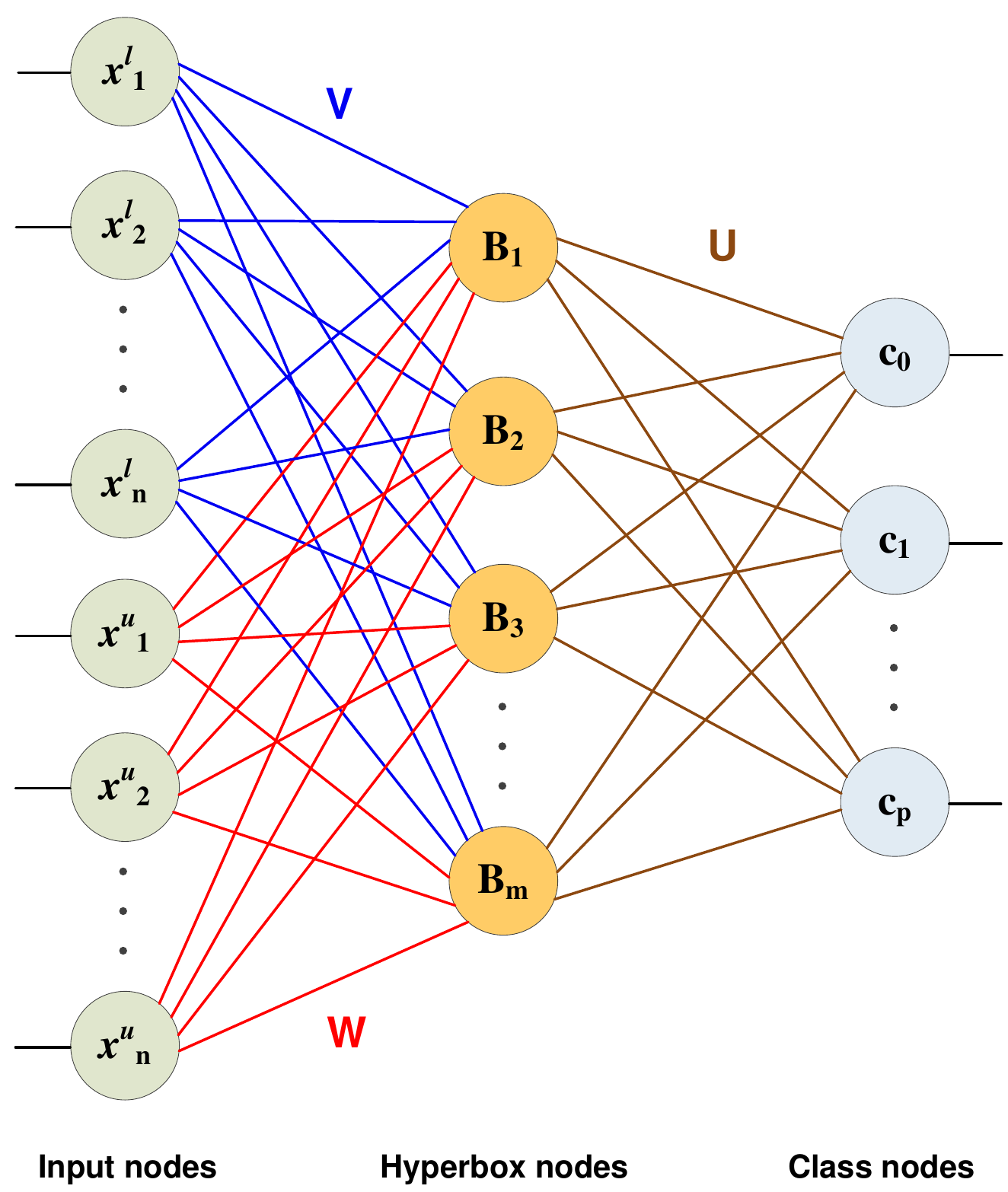}
	\caption{The architecture of GFMM neural network.}
	\label{Fig.1}
\end{figure}

General fuzzy min-max (GFMM) neural network \cite{Gabrys00} is a generalization and extension of the fuzzy min-max neural network (FMNN) \cite{Simpson92}. It combines both classification and clustering in a unified framework and can deal with both fuzzy and crisp input samples. The architecture of the general fuzzy min-max neural network comprises three layers, i.e., an input layer, a hyperbox fuzzy set layer, and an output (class) layer, shown in Fig. \ref{Fig.1}.

The input layer contains $ 2 \cdot n $ nodes, where $ n $ is the number of dimensions of the problem, to fit with the input sample $ X = [X^l, X^u]$, determined within the n-dimensional unit cube $\mathbf{I}^n$. The first $ n $ nodes in the input layer are connected to $ m $ nodes of the second layer, which contains hyperbox fuzzy sets, by the minimum points matrix \textbf{V}. The remaining $ n $ nodes are linked to $ m $ nodes of the second layer by the maximum points matrix \textbf{W}. The values of two matrices \textbf{V} and \textbf{W} are adjusted during the learning process. Each hyperbox $B_i$ is defined by an ordered set: $B_i = \{X, V_i, W_i, b_i(X, V_i, W_i)\}$, where $V_i \subset \mathbf{V}, W_i \subset \mathbf{W}$ are minimum and maximum points of hyperbox $B_i$ respectively, $b_i$ is the membership value of hyperbox $ B_i $ in the second layer, and it is also the transfer function between input nodes in the first layer and hyperbox nodes in the second layer. The membership function $b_i$ is computed using \eqref{eqmem} \cite{Gabrys00}.

\begin{equation}
\label{eqmem}
\begin{aligned}
b_i(X, V_i, W_i) = \min \limits_{j = 1 \ldots n} (\min (&[1 - f(x_{j}^u - w_{ij}, \gamma_j)], \\ 
& [1 - f(v_{ij} - x_{j}^l, \gamma_j)]))
\end{aligned}
\end{equation}
where $ f(r, \gamma) = \begin{cases} 
1, & \mbox{if } r\gamma > 1 \\
r\gamma, & \mbox{if } 0 \leq r\gamma \leq 1 \\
0, & \mbox{if } r\gamma < 0 \\
\end{cases}
$ is the threshold function and $ \gamma = [ \gamma_1,\ldots, \gamma_n ]$ is a sensitivity parameter regulating the speed of decrease of the membership values.

Hyperboxes in the middle layer are fully connected to the third-layer nodes by a binary valued matrix \textbf{U}. The elements in the matrix \textbf{U} are computed as follows:
\begin{equation}
u_{ij} = \begin{cases}
1, \quad \mbox{if hyperbox $B_i$ represents class $ c_j $} \\
0, \quad \mbox{otherwise}
\end{cases}
\end{equation}
where $ B_i $ is the hyperbox of the second layer, and $ c_j $ is the $ j^{th} $ node in the third layer. Output of each node $ c_j $ in the third layer is a membership degree to which the input pattern $ X $ fits within the class $ j $. The transfer function of each node $ c_j $ in $ p + 1 $ nodes belonging to the third layer is computed as:

\begin{equation}
\label{eq2}
c_j = \max \limits_{i = 1}^m {b_i \cdot u_{ij}}
\end{equation}
Node $ c_0 $ is connected to all unlabeled hyperboxes of the middle layer. The values of nodes in the output layer can be fuzzy if they are computed directly from \eqref{eq2} or crisp in the case that the node with the largest value is assigned to 1 and the others get a value of zero \cite{Gabrys00}.

The incremental learning algorithm for the GFMM neural network to adjust the values of two matrices \textbf{V} and \textbf{W} includes four steps, i.e., initialization, expansion, hyperbox overlap test, and contraction, in which the last three steps are repeated. In the initialization stage, each hyperbox $ B_i $ is initialized with $ V_i = 1 $ and $ W_i = 0 $. For each input sample $ X $, the algorithm finds the hyperbox $ B_i $ with the highest membership value representing the same class as $ X $ to verify two expansion conditions, i.e., maximum allowable hyperbox size and class label compatibility as shown in the supplemental file. If both criteria are met, the selected hyperbox is expanded. If no hyperbox meets the expansion conditions, a new hyperbox is created to accommodate the input data. Otherwise, if the hyperbox $ B_i $ is expanded in the prior step, it will be checked for an overlap with other hyperboxes $ B_k $ as follows. If the class of $ B_i $ is equal to zero, then the hyperbox $ B_i $ must be checked for overlap with all existing hyperboxes; otherwise, the overlap checking is only performed between $ B_i $ and hyperboxes $ B_k $ representing other classes. If the overlap occurs, a contraction process is carried out to remove the overlapping area by adjusting the sizes of hyperboxes according to the dimension with the smallest overlapping value. Four cases of the overlap checking and contraction procedures were presented in detail in the supplemental file and can be found in \cite{Gabrys00}.

\section{Proposed Methods}\label{method}
\subsection{Overview}
The learning process of the proposed method consists of two phases. The first phase is to rapidly construct small-sized hyperbox fuzzy sets from similar input data points. This phase is performed in parallel on training data segments. The data in each fragment can be organized according to two modes. The first way is called heterogeneous mode, which uses the data order read from the input file. The second mode is homogeneous, in which the data are sorted according to groups; each group contains data from the same class. The main purpose of the second phase is to decrease the complexity of the model by reconstructing phase-1 hyperboxes with a higher abstraction level.

\begin{figure*} [!ht]
	\centering
	\includegraphics[width=0.8\linewidth]{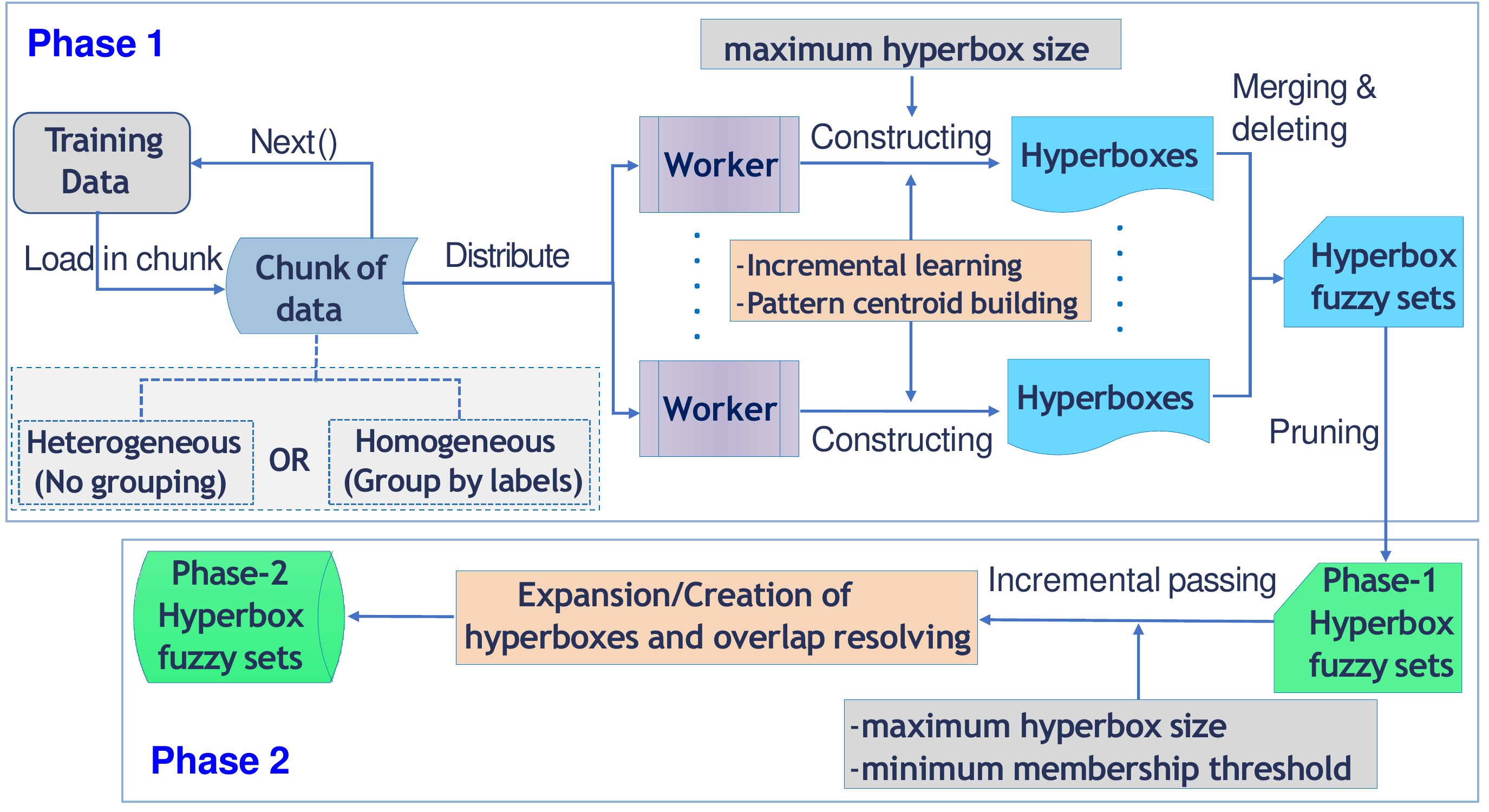}
	\caption{Pipeline of the training process of the proposed method.}
	\label{Fig.2}
\end{figure*}

In the first step of the training process, the input samples are split into disjoint sets and are then distributed to different computational workers. On each worker, we build an independent general fuzzy min-max neural network. When all training samples are handled, all created hyperboxes at different workers are merged to form a single model. Hyperboxes completely included in other hyperboxes representing the same class are eliminated to reduce the redundancy and complexity of the final model. After combining hyperboxes, the pruning procedure needs to be applied to eliminate noise and low-quality hyperboxes. The resulting hyperboxes are called phase-1 hyperboxes.

However, phase-1 hyperboxes have small sizes, so the complexity of the system can be high. As a result, all these hyperboxes are put through phase-2 of the granulation process with a gradual increase in the maximum hyperbox sizes. At a larger value of the maximum hyperbox size, hyperboxes at a low level of abstraction will be reconstructed with a higher data abstraction degree. Previously generated hyperboxes are fetched one at a time, and they are aggregated with newly constructed hyperboxes at the current granular representation level based on a similarity threshold of the membership degree. This process is repeated for each input value of the maximum hyperbox sizes. The whole process of the proposed method is shown in Fig. \ref{Fig.2}. Based on the classification error of resulting classifiers on the validation set, one can select an appropriate predictor satisfying both simplicity and precision. The following part provides the core concepts for both phases of our proposed method in a form of mathematical descriptions. The details of Algorithm 1 for the phase 1 and Algorithm 2 corresponding to the phase 2 as well as their implementation aspects are shown in the supplemental material. The readers can refer to this document to find more about the free text descriptions, pseudo-codes, and implementation pipeline of the algorithms.

\subsection{Formal Description}
Consider a training set of $N_{Tr}$ data vectors, $\mathbf{X}^{(Tr)} = \{X_i^{(Tr)}: X_i^{(Tr)} \in \mathbb{R}^n, i = 1,\ldots,N_{Tr} \}$, and the corresponding classes, $\mathbf{C}^{(Tr)} = \{c_i^{(Tr)}: c_i^{(Tr)} \in \mathbb{N}, i = 1,\ldots,N_{Tr} \}$; a validation set of $N_{V}$ data vectors, $\mathbf{X}^{(V)} = \{X_i^{(V)}: X_i^{(V)} \in \mathbb{R}^n, i = 1,\ldots,N_{V} \}$, and the corresponding classes, $\mathbf{C}^{(V)} = \{c_i^{(V)}: c_i^{(V)} \in \mathbb{N}, i = 1,\ldots,N_{V} \}$. The details of the method are formally described as follows.

\subsubsection{Phase 1}\hfill

Let $n_w$ be the number of workers to execute the hyperbox construction process in parallel. Let $\mathbb{F}_j(\mathbf{X}_j^{(Tr)}, \mathbf{C}_j^{(Tr)}, \theta_0)$ be the procedure to construct hyperboxes on the $j^{th}$ worker with maximum hyperbox size $\theta_0$ using training data $\{\mathbf{X}_j^{(Tr)}, \mathbf{C}_j^{(Tr)}\}$. Procedure $\mathbb{F}_j$ is a modified fuzzy min-max neural network model which only creates new hyperboxes or expands existing hyperboxes. It accepts the overlapping regions among hyperboxes representing different classes, because we expect to capture rapidly similar samples and group them into specific clusters by small-sized hyperboxes without spending much time on computationally expensive hyperbox overlap test and resolving steps. Instead, each hyperbox $ B_i $ is added a centroid $ G_i $ of patterns contained in that hyperbox and a counter $ N_i $ to store the number of data samples covered by it in addition to maximum and minimum points. This information is used to classify data points located in the overlapping regions. When a new pattern $ X $ is presented to the classifier, the operation of building the pattern centroid for each hyperbox (line 12 and line 15 in Algorithm 1) is performed according to \eqref{centroid}.

\begin{equation}
\label{centroid}
\footnotesize
G_i^{new} = \cfrac{N_i \cdot G_i^{old} + X}{N_i + 1}
\end{equation}
where $ G_i $ is the sample centroid of the hyperbox $ B_i $, $ N_i $ is the number of current samples included in the $ B_i $. Next, the number of samples is updated: $ N_i = N_i + 1 $. It is noted that $ G_i $ is the same as the first pattern covered by the hyperbox $ B_i $ when $ B_i $ is newly created.

After the process of building hyperboxes in all workers finishes, merging step is conducted (lines 19-23 in Algorithm 1), and it is mathematically represented as:

\begin{equation}
\label{merging}
\footnotesize
\mathbf{M} = \{B_i | B_i \in \bigcup\limits_{j = 1}^{n_w} \mathbb{F}_j(\mathbf{X}_j^{(Tr)}, \mathbf{C}_j^{(Tr)}, \theta_0) \}
\end{equation}
where $\mathbf{M}$ is the model after performing the merging procedure. It is noted that hyperboxes contained in the larger hyperboxes representing the same class are eliminated (line 24 in Algorithm 1) and the centroids of larger hyperboxes are updated using \eqref{updatecentroid}.

\begin{equation}
\label{updatecentroid}
\footnotesize
G_1^{new} = \cfrac{N_1 \cdot G_1^{old} + N_2 \cdot G_2^{old}}{N_1 + N_2}
\end{equation}
where $ G_1 $ and $ N_1 $ are the centroid and the number of samples of the larger sized hyperbox, $ G_2 $ and $ N_2 $ are the centroid and the number of samples of the smaller sized hyperbox. The number of samples in the larger sized hyperbox is also updated: $ N_1 = N_1 + N_2 $. This whole process is similar to the construction of an ensemble classifier at the model level shown in \cite{Gabrys02}.

Pruning step is performed after merging hyperboxes to remove noise and low-quality hyperboxes (line 26 in Algorithm 1). Mathematically, it is defined as:
\begin{equation}
\label{pruning}
\footnotesize
\mathbf{H}_0 = \begin{cases}
\mathbf{M}_1 = \mathbf{M} \setminus \{B_k | \mathcal{A}_k < \alpha\ \vee \mathcal{A}_k = Nil\}  \mbox{, if } \mathcal{E}_V(\mathbf{M}_1) \le \mathcal{E}_V(\mathbf{M}_2) \\
\mathbf{M}_2 = \mathbf{M} \setminus \{B_k | \mathcal{A}_k < \alpha\}  \mbox{, otherwise}
\end{cases}
\end{equation}
where $\mathbf{H}_0$ is the final model of stage 1 after applying the pruning operation, $ \mathcal{E}_V(\mathbf{M}_i) $ is the classification error of the model $\mathbf{M}_i$ on the validation set $\{\mathbf{X}^{(V)}, \mathbf{C}^{(V)}\}$, $\alpha$ is the minimum accuracy of each hyperbox to be retained, and $ \mathcal{A}_k $ is the predictive accuracy of hyperbox $B_k \in \mathbf{M}$ on the validation set defined as follows:

\begin{equation}
\footnotesize
\mathcal{A}_k = \cfrac{\sum \limits_{j = 1}^{S_k} \mathcal{R}_{kj}}{\sum \limits_{j = 1}^{S_k} (\mathcal{R}_{kj} + \mathcal{I}_{kj})}
\end{equation}
where $S_k$ is the number of validation samples classified by hyperbox $B_k$, $\mathcal{R}_k$ is the number of samples predicted correctly by $B_k$, and $\mathcal{I}_k$ is the number of incorrect predicted samples. If $S_k =0$, then $A_k = Nil$.

The classification step of unseen samples using model $\mathbf{H}_0$ is performed in the same way as in the GFMM neural network with an exception of the case of many winning hyperboxes with the same maximum membership value. In such a case, we compute the Euclidean distance from the input sample $X$ to centroids $G_i$ of winning hyperboxes $B_i$ using \eqref{dis}. If the input sample is a hyperbox, $X$ is the coordinate of the center point of that hyperbox.

\begin{equation}
\label{dis}
\footnotesize
d(X, G_i) = \sqrt{\sum \limits_{j = 1}^n (x_j - G_{ij})^2}
\end{equation}
The input pattern is then classified to the hyperbox $B_i$ with the minimum value of $d(X, G_i)$.

\subsubsection{Phase 2}\hfill

Unlike phase 1, the input data in this phase are hyperboxes generated in the previous step. The purpose of stage 2 is to reduce the complexity of the model by aggregating hyperboxes created at a higher resolution level of granular data representations. At the high level of data abstraction, the confusion among hyperboxes representing different classes needs to be removed. Therefore, the overlapping regions formed in phase 1 have to be resolved, and there is no overlap allowed in this phase. Phase 2 can be mathematically represented as:

\begin{equation}
\label{phase2}
\small
\mathbf{H}_H(\Theta, m_s) = \{\mathbf{H}_i | \mathbf{H}_i = \mathbb{G}(\mathbf{H}_{i-1}, \theta_i, m_s) \}, \forall i \in [1, |\Theta|], \theta_i \in \Theta
\end{equation} 
where $\mathbf{H}_H$ is a list of models $\mathbf{H}_i$ constructed through different levels of granularity represented by maximum hyperbox sizes $\theta_i$, $\Theta$ is a list of maximum hyperbox sizes, $|\Theta| $ is the cardinality of $\Theta$, $m_s$ is the minimum membership degree of two aggregated hyperboxes, and $\mathbb{G}$ is a procedure to construct the models in phase 2 (it uses the model at previous step as input), $\mathbf{H}_0$ is the model in phase 1. The aggregation rule of hyperboxes, $\mathbb{G}$, is described as follows:

For each input hyperbox $ B_h $ in $\mathbf{H}_{i -1}$, the membership values between $ B_h $ and all existing hyperboxes with the same class as $B_h$ in $\mathbf{H}_{i}$ are computed. We select the winner hyperbox with maximum membership degree with respect to $B_h$, denoted $B_k$, to aggregate with $B_h$. The following constraints are verified before conducting the aggregation:

\begin{itemize}
	\item Maximum hyperbox size:\\ 
	\begin{equation}
	\max(w_{hj}, w_{kj}) - \min(v_{hj}, w_{kj}) \leq \theta_i, \quad \forall j \in [1, n]
	\end{equation}
	\item The minimum membership degree:
	\begin{equation}
	b(B_h, B_k) \geq m_s
	\end{equation}
	\item Overlap test. New hyperbox aggregated from $B_h$ and $B_k$ does not overlap with any existing hyperboxes in $\mathbf{H}_i$ belonging to other classes
\end{itemize}

If hyperbox $B_k$ has not met all of the above conditions, the hyperbox with the next highest membership value is selected and the process is repeated until the aggregation step occurs or no hyperbox candidate is left. If the input hyperbox cannot be mergered with existing hyperboxes in $\mathbf{H}_i$, it will be directly inserted into the current list of hyperboxes in $\mathbf{H}_i$. After that, the overlap test operation between the newly inserted hyperbox and hyperboxes in $\mathbf{H}_i$ representing other classes is performed, and then the contraction process will be executed to resolve overlapping regions. The algorithm is iterated for all input hyperboxes in $\mathbf{H}_{i - 1}$.

The classification process for unseen patterns using the hyperboxes in phase 2 is realized as in the GFMM neural network. A detailed decription of the implementation steps for the proposed method can be found in the supplemental material.

\subsection{Missing Value Handling}
The proposed method can deal with missing values since it inherits this characteristic from the general fuzzy min-max neural network as shown in \cite{Gabrys02b}. A missing feature $ x_j $ is assumed to be able to receive values from the whole range, and it is presented by a real-valued interval as follows: $ x_{j}^l = 1, x_{j}^u = 0 $. By this initialization, the membership value associated with the missing value will be one, and thus the missing value does not cause any influence on the membership function of the hyperbox. During the training process, only observed features are employed for the update of hyperbox minimum and maximum vertices while missing variables are disregarded automatically. For the overlap test procedure in phase 2, only the hyperboxes which satisfy $ v_{ij} \leq w_{ij} $ for all dimensions $ j \in [1, n] $ are verified for the undesired overlapping areas. The second change is related to the way of calculating the membership value for the process of hyperbox selection or classification step of an input sample with missing values. Some hyperboxes' dimensions have not been set, so the membership function shown in \eqref{eqmem} is changed to $ b_i(X, \min(V_i, W_i), \max(W_i, V_i)) $. With the use of this method, the training data uncertainty is represented in the classifier model.

\section{Experiments}\label{experiment}
Marcia et al. \cite{Macia13} argued that data set selection poses a considerable impact on conclusions of the accuracy of learners, and then the authors advocated for considering properties of the datasets in experiments. They indicated the importance of employing artificial data sets constructed based on previously defined characteristics. In these experiments, therefore, we considered two types of synthetic datasets with linear and non-linear class boundaries. We also changed the number of features, the number of samples, and the number of classes for synthetic datasets to assess the variability in the performance of the proposed method. In practical applications, the data are usually not ideal as well as not following a standard distribution rule and including noisy data. Therefore, we also carried out experiments on real datasets with diversity in the numbers of samples, features, and classes.

For medium-sized real datasets such as \textit{Letter}, \textit{Magic}, \textit{White wine quality}, and \textit{Default of credit card clients}, the density-preserving sampling (DPS) method \cite{Budka13} was used to separate the original datasets into training, validation, and test sets. For large-sized datasets, we used the hold-out method for splitting datasets, which is the simplest and the least computationally expensive approach to assessing the performance of classifiers because more advanced resampling approaches are not essential for large amounts of data \cite{Budka13}. The classification model is then trained on the training dataset. The validation set is used for the pruning step and evaluating the performance of the constructed classifier aiming to select a suitable model. The testing set is employed to assess the efficiency of the model on unseen data. 

The experiments aim to answer the following questions:

\begin{itemize}
	\item How is the classification accuracy of the predictor using multi-resolution hierarchical granular representations improved in comparison to the model using a fixed granulation value?
	
	\item How good is the performance of the proposed method compared with other types of fuzzy min-max neural networks and popular algorithms based on other data representations such as support vector machines, Naive Bayes, and decision tree?
	
	\item Whether we can obtain a classifier with high accuracy at high abstraction levels of granular representations?
	
	\item Whether we can rely on the performance of the model on validation sets to select a good model for unseen data, which satisfies both simplicity and accuracy?
	
	\item How good is the ability of handling missing values in datasets without requiring data imputation?
	
	\item How critical are the roles of the pruning process and the use of sample centroids?
\end{itemize}

The limitation of runtime for each execution is seven days. If an execution does not finish within seven days, it will be terminated, and the result is reported as N/A. In the experiments, we set up parameters as follows: $ n_w = 4, \alpha = 0.5, m_s = 0.4, \gamma = 1$ because they gave the best results on a set of preliminary tests with validation sets for the parameter selection. All datasets are normalized to the range of [0, 1] because of the characteristic of the fuzzy min-max neural networks. Most of the datasets except the \textit{SUSY} datasets utilized the value of 0.1 for $ \theta_0 $ in phase 1, and $\Theta = \{0.2, 0.3, 0.4, 0.5, 0.6 \}$ for different levels of granularity in phase 2. For the \textit{SUSY} dataset, due to the complexity and limitation of runtime for the proposed method and other compared types of fuzzy min-max neural networks, the $ \theta_0 = 0.3 $ was used for phase 1, and $ \Theta = \{0.4, 0.5, 0.6\} $ was employed for phase 2. For Naive Bayes, we used Gaussian Naive Bayes (GNB) algorithm for classification. The radial basis function (RBF) was used as a kernel function for the support vector machines (SVM). We used the default setting parameters in the \textit{scikit-learn} library for Gaussian Naive Bayes, SVM, and decision tree (DT) in the experiments. The performance of the proposed method was also compared to other types of fuzzy min-max neural networks such as the original fuzzy min-max neural network (FMNN) \cite{Simpson92}, the enhanced fuzzy min-max neural network (EFMNN) \cite{Mohammed15}, the enhanced fuzzy min-max neural network with a K-nearest hyperbox expansion rule (KNEFMNN) \cite{Mohammed17}, and the general fuzzy min-max neural network (GFMMNN) \cite{Gabrys00}. These types of fuzzy min-max neural networks used the same pruning procedure as our proposed method.

Synthetic datasets in our experiments were generated by using Gaussian distribution functions, so Gaussian Naive Bayes and SVM with RBF kernel which use Gaussian distribution assumptions to classify data will achieve nearly optimal error rates because they match perfectly with underlying data distribution. Meanwhile, fuzzy min-max classifiers employ the hyperboxes to cover the input data, thus they are not an optimal representation for underlying data. Therefore, the accuracy of hyperbox-based classifiers on synthetic datasets cannot outperform the predictive accuracy of Gaussian NB or SVM with RBF kernel. However, Gaussian NB is a linear classifier, and thus, it only outputs highly accurate predictive results for datasets with linear decision boundary. In contrast, decision tree, fuzzy min-max neural networks, and SVM with RBF kernel are universal approximators, and they can deal effectively with both linear and non-linear classification problems.

All experiments were conducted on the computer with Xeon 6150 2.7GHz CPU and 180GB RAM. We repeated each experiment five times to compute the average training time. The accuracy of types of fuzzy min-max neural networks remains the same through different iterations because they only depend on the data presentation order and we kept the same order of training samples during the experiments.

\subsection{Performance of the Proposed Method on Synthetic Datasets}

The first experiment was conducted on the synthetic datasets with the linear or non-linear boundary between classes. For each synthetic dataset, we generated a testing set containing 100,000 samples and a validation set with 10,000 instances using the same probability density function as the training sets. 

\subsubsection{Linear Boundary Datasets}~\\
\textbf{\textit{Increase the number of samples:}}

\begin{figure*}
	\begin{subfloat}[10K samples]{
			\includegraphics[width=0.3\textwidth]{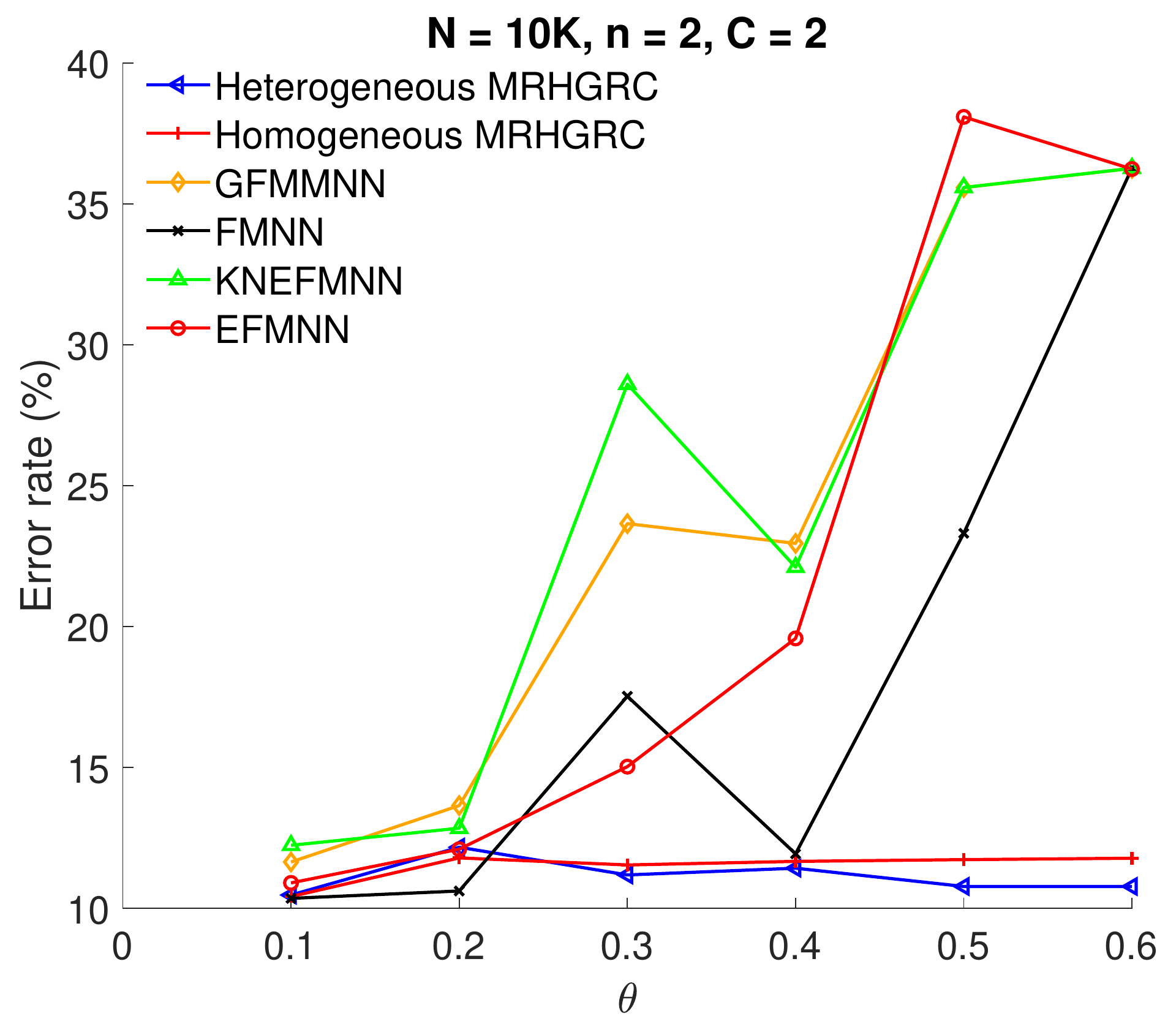}}
	\end{subfloat}
	\hfill%
	\begin{subfloat}[1M samples]{
			\includegraphics[width=0.3\textwidth]{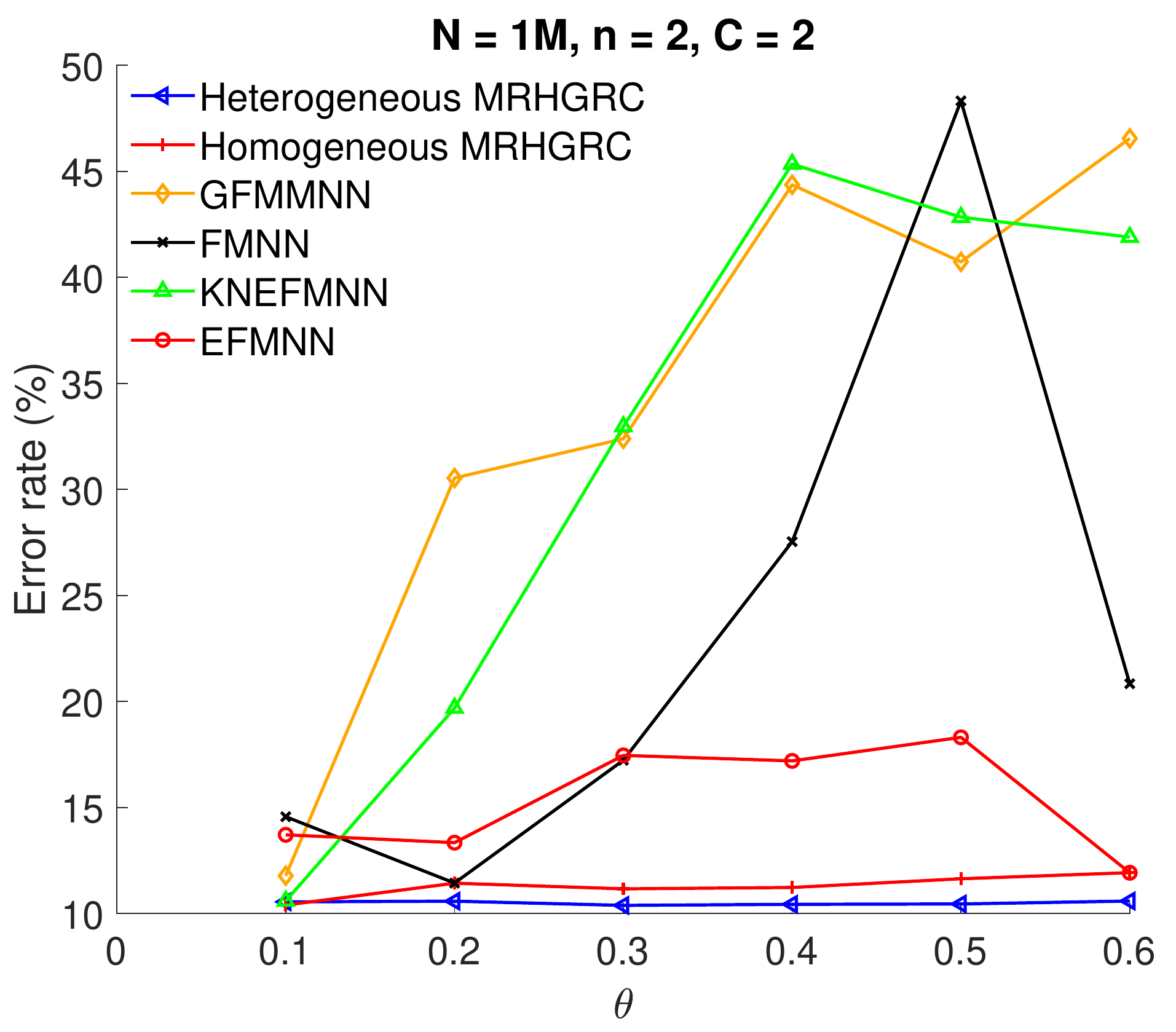}}
	\end{subfloat}
	\hfill%
	\begin{subfloat}[5M samples]{
			\includegraphics[width=0.3\textwidth]{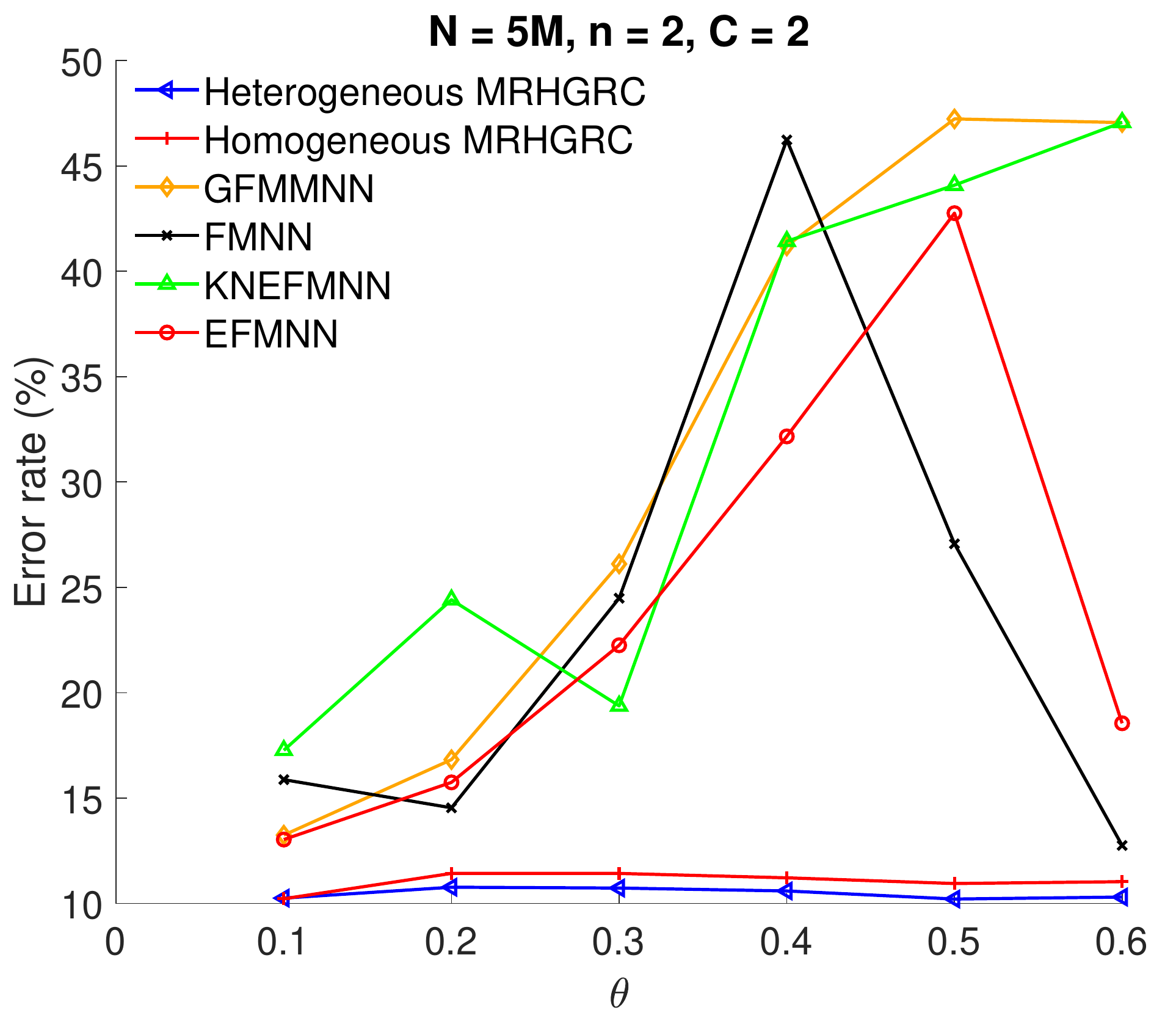}}
	\end{subfloat}
	\hfill%
	\caption{The error rate of classifiers on synthetic linear boundary datasets with the different number of samples.}
	\label{Fig.3}
\end{figure*}

We kept both the number of dimensions $ n = 2 $ and the number of classes $ C = 2 $ the same, and only the number of samples was changed to evaluate the impact of the number of patterns on the performance of classifiers. We used Gaussian distribution to construct synthetic datasets as described in \cite{Fukunaga90}. The means of the Gaussians of two classes are given as follows: $ \mu_1 = [0, 0]^T, \mu_2 = [2.56, 0]^T $, and the covariance matrices are as follows:
\begin{equation*}
\scriptsize
\Sigma_1 = \Sigma_2 = \left[ \begin{array}{cc}
1&0\\
0&1\\
\end{array}
\right]
\end{equation*}
With the use of these configuration parameters, training sets with different sizes (10K, 1M, and 5M samples) were formed. Fukunaga \cite{Fukunaga90} indicated that the general Bayes error of the datasets formed from these settings is around 10\%. We generated an equal number of samples for each class to remove the impact of imbalanced class property on the performance of classifiers. Fig. \ref{Fig.3} shows the change in the error rates of different fuzzy min-max classifiers on the testing synthetic linear boundary datasets with the different numbers of training patterns when the level of granularity ($\theta$) changes. The other fuzzy min-max neural networks used the fixed value of $ \theta $ to construct the model, while our method builds the model starting from the defined lowest value of $ \theta $ (phase 1) to the current threshold. For example, the model at $ \theta = 0.3 $ in our proposed method is constructed with $ \theta = 0.1 $, $ \theta = 0.2 $, and $ \theta = 0.3 $.

\begin{table} [!ht]
	\centering
	\caption{The Lowest Error Rates and Training Time of Classifiers on Synthetic Linear Boundary Datasets With Different Number of Samples ($ n = 2, C = 2 $)}
	\label{Table.1}
	{\scriptsize
		\begin{tabular}{|c|l|r|r|c|c|r|}
			\hline
			\bfseries N            & \bfseries Algorithm            & \bfseries $\min E_{V}$ & \bfseries $\min E_{T}$ & \bfseries $\theta_{V}$ & \bfseries $\theta_{T}$ & \bfseries Time (s) \\ 
			\hline \hline
			\multirow{9}{*}{10K}  & He-MRHGRC            & 10.25           & 10.467           & 0.1            & 0.1             & 1.1378            \\ \cline{2-7} 
			& Ho-MRHGRC            & 10.1            & 10.413           & 0.1            & 0.1             & 1.3215            \\ \cline{2-7} 
			& GFMM                 & 11.54           & 11.639           & 0.1            & 0.1             & 8.6718            \\ \cline{2-7} 
			& FMNN                 & 10.05           & 10.349           & 0.1            & 0.1             & 46.4789           \\ \cline{2-7} 
			& KNEFMNN              & 12.07           & 12.232           & 0.1            & 0.1             & 9.4459            \\ \cline{2-7} 
			& EFMNN                & 10.44           & 10.897           & 0.1            & 0.1             & 48.9892           \\ \cline{2-7} 
			& GNB                  & \textbf{9.85}            & \textbf{9.964}            & -              & -               & 0.5218            \\ \cline{2-7} 
			& SVM                  & 9.91            & 9.983            & -              & -               & 1.5468         \\ \cline{2-7} 
			& DT                   & 15.33           & 14.861           & -              & -               & 0.5405         \\ \hline
			\multirow{9}{*}{1M}   & He-MRHGRC            & 10.31           & 10.386           & 0.3            & 0.3             & 20.0677           \\ \cline{2-7} 
			& Ho-MRHGRC            & 10.24           & 10.401           & 0.1            & 0.1             & 16.0169           \\ \cline{2-7} 
			& GFMM                 & 11.47           & 11.783           & 0.1            & 0.1             & 405.4642          \\ \cline{2-7} 
			& FMNN                 & 10.98           & 11.439           & 0.2            & 0.2             & 13163.1404        \\ \cline{2-7} 
			& KNEFMNN              & 10.36           & 10.594           & 0.1            & 0.1             & 413.8296          \\ \cline{2-7} 
			& EFMNN                & 11.61           & 11.923           & 0.6            & 0.6             & 10845.1280       \\ \cline{2-7} 
			& GNB                  & 9.87            & \textbf{9.972}            & -              & -               & 5.0133         \\ \cline{2-7} 
			& SVM                  & \textbf{9.86}            & 9.978            & -              & -               & 21798.2803       \\ \cline{2-7} 
			& DT                   & 14.873          & 14.682           & -              & -               & 9.9318         \\ \hline
			\multirow{9}{*}{5M}   & He-MRHGRC            & 10.11           & 10.208           & 0.5            & 0.5             & 101.9312          \\ \cline{2-7} 
			& Ho-MRHGRC            & 10.04           & 10.222           & 0.1            & 0.1             & 75.2254           \\ \cline{2-7} 
			& GFMM                 & 13.14           & 13.243           & 0.1            & 0.1             & 1949.2138         \\ \cline{2-7} 
			& FMNN                 & 12.68           & 12.751           & 0.6            & 0.6             & 92004.7253        \\ \cline{2-7} 
			& KNEFMNN              & 17.31           & 17.267           & 0.1            & 0.1             & 1402.1173         \\ \cline{2-7} 
			& EFMNN                & 12.89           & 13.032           & 0.1            & 0.1             & 41888.5296        \\ \cline{2-7} 
			& GNB                  & \textbf{9.88}            & \textbf{9.976}            & -              & -               & 22.9343        \\ \cline{2-7} 
			& SVM                  & N/A             & N/A              & -              & -               & N/A               \\ \cline{2-7} 
			& DT                   & 15.253          & 14.692           & -              & -               & 70.2041        \\ \hline
		\end{tabular}
	}
\end{table}

It can be seen from Fig. \ref{Fig.3} that the error rates of our method are lower than those of other fuzzy min-max classifiers, especially in high abstraction levels of granular representations. At high levels of abstraction (corresponding to high values of $\theta$), the error rates of other classification models are relatively high, while our proposed classifier still maintains the low error rate, just a little higher than the error at a high-resolution level of granular data. The lowest error rates of the different classifiers on validation ($ E_V $) and testing ($ E_T $) sets, as well as total training time for six levels of abstraction, are shown in Table \ref{Table.1}. Best results are highlighted in bold in each table. The training time reported in this paper consists of time for reading training files and model construction.

We can see that the accuracy of our method on unseen data using the heterogeneous data distribution (He-MRHGRC) regularly outperforms the accuracy of the classifier built based on the homogeneous data distribution (Ho-MRHGRC) using large-sized training sets. It is also observed that our method is less affected by overfitting when increasing the number of training samples while keeping the same testing set. For other types of fuzzy min-max neural networks, their error rates frequently increase with the increase in training size because of overfitting. The total training time of our algorithm is faster than that of other types of fuzzy min-max classifiers since our proposed method executes the hyperbox building process at the lowest value of $ \theta $ in parallel, and we accept the overlapping areas among hyperboxes representing different classes to rapidly capture the characteristics of sample points locating near each other. The hyperbox overlap resolving step is only performed at higher abstraction levels with a smaller number of input hyperboxes.

Our proposed method also achieves better classification accuracy compared to the decision tree, but it cannot overcome the support vector machines and Gaussian Naive Bayes methods on synthetic linear boundary datasets. However, the training time of the support vector machines on large-sized datasets is costly, even becomes unacceptable on training sets with millions of patterns. The synthetic datasets were constructed based on the Gaussian distribution, so the Gaussian Naive Bayes method can reach the minimum error rate, but our approach can also obtain the error rates relatively near these optimal error values. We can observe that the best performance of the He-MRHGRC attains at quite high abstraction levels of granular representations because some noisy hyperboxes at high levels of granularity are eliminated at lower granulation levels. These results demonstrate the efficiency and scalability of our proposed approach.

\textbf{\textit{Increase the number of classes:}}

The purpose of the experiment in this subsection is to evaluate the performance of the proposed method on multi-class datasets. We kept the number of dimensions $ n = 2 $, the number of samples $ N = 10,000$, and only changed the number of classes to form synthetic multi-class datasets with $ C \in \{2, 4, 16\} $. The covariance matrices stay the same as in the case of changing the number of samples.

\begin{figure*}
	\begin{subfloat}[2 classes]{
			\includegraphics[width=0.3\textwidth]{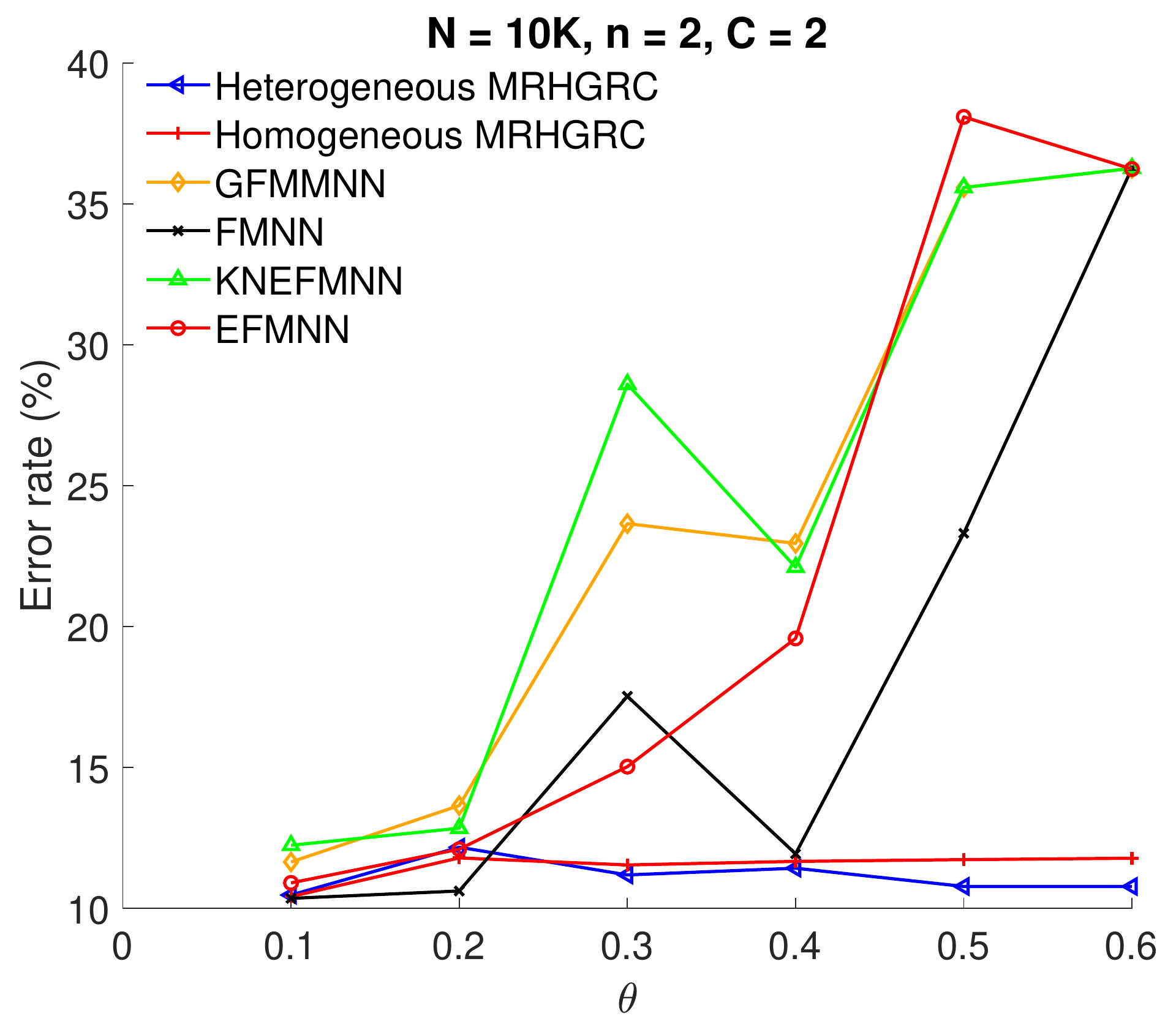}}
	\end{subfloat}
	\hfill%
	\begin{subfloat}[4 classes]{
			\includegraphics[width=0.3\textwidth]{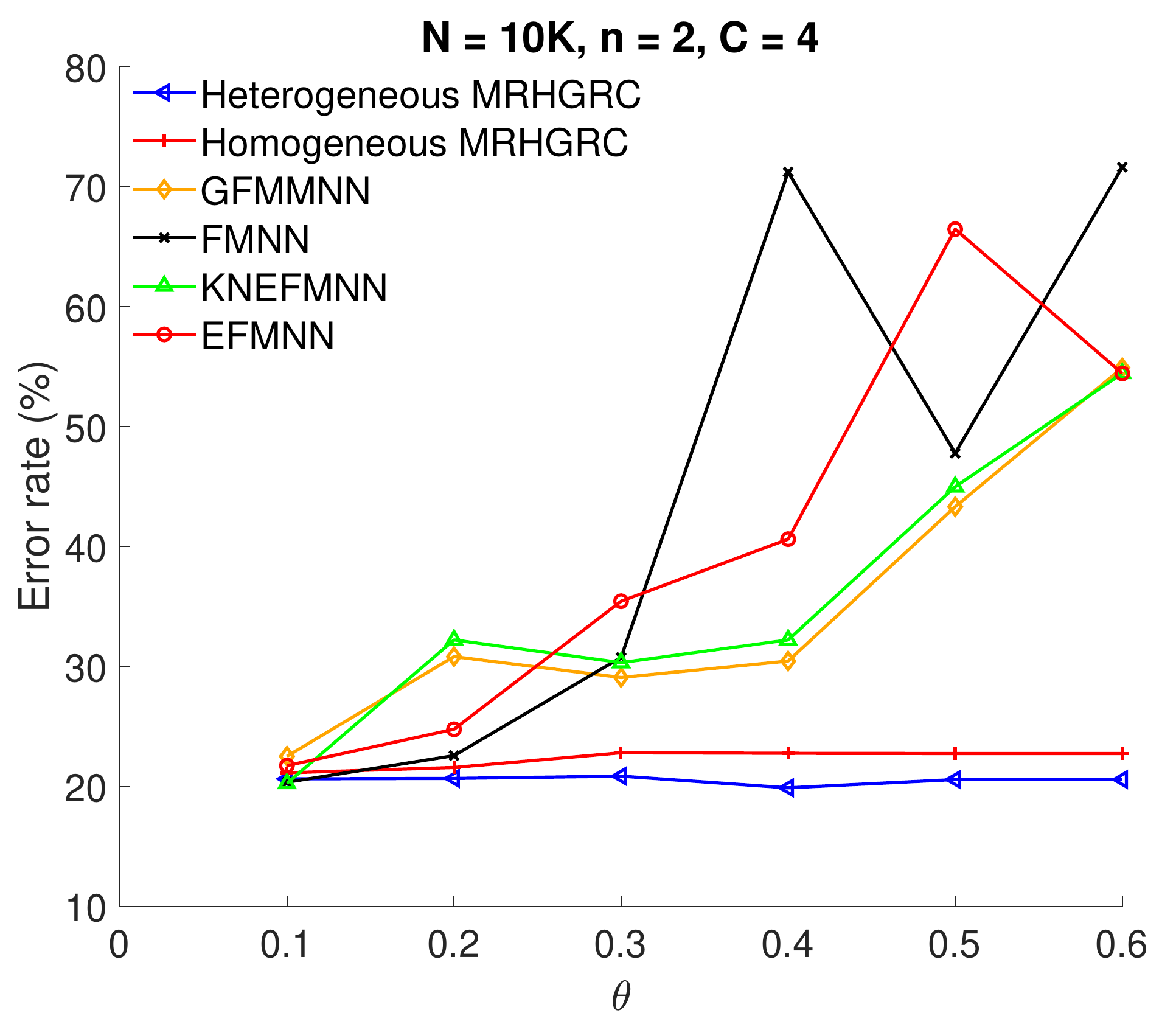}}
	\end{subfloat}
	\hfill%
	\begin{subfloat}[16 classes]{
			\includegraphics[width=0.3\textwidth]{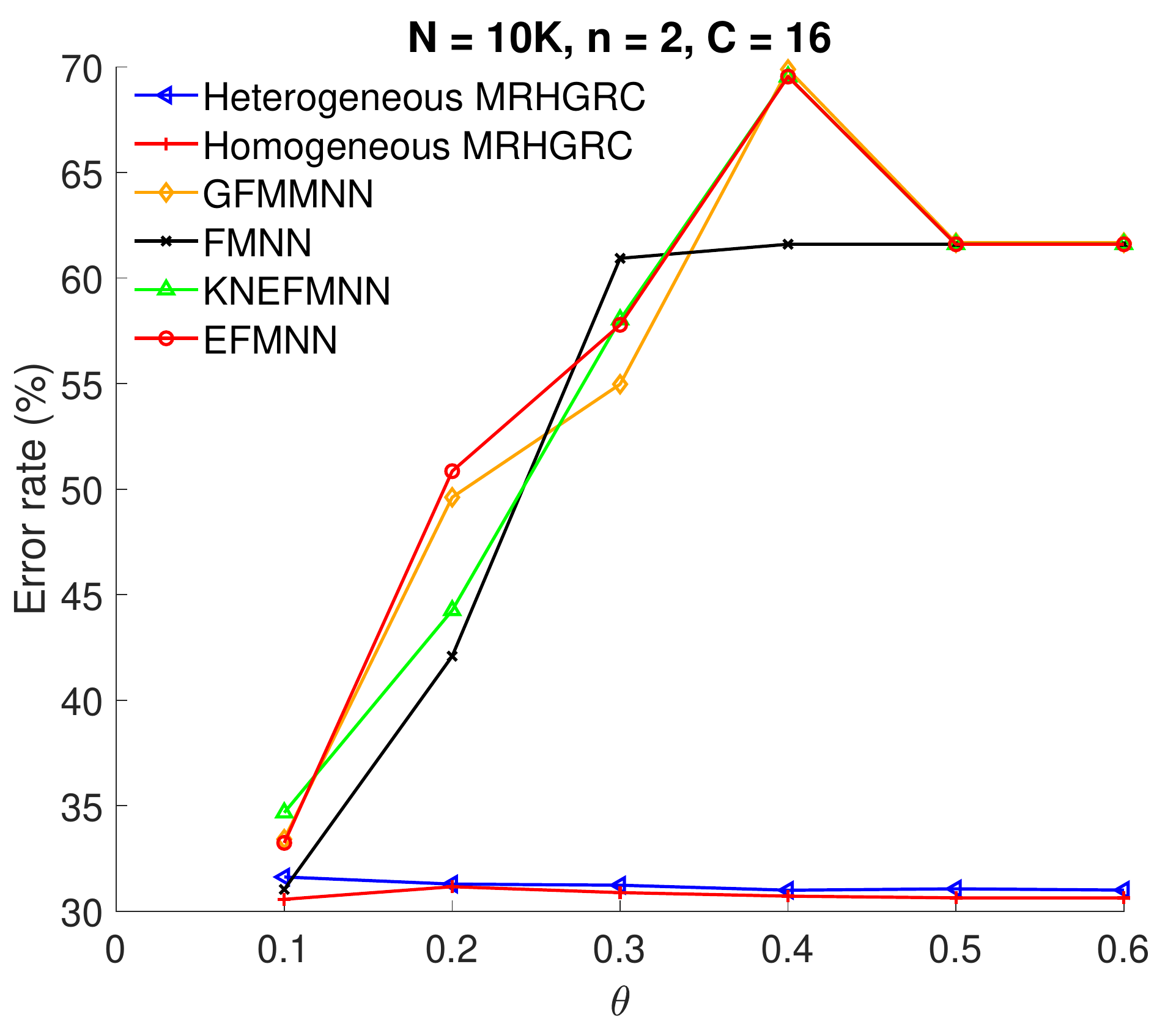}}
	\end{subfloat}
	\hfill%
	\caption{The error rate of classifiers on synthetic linear boundary datasets with the different number of classes.}
	\label{Fig.4}
\end{figure*}

\begin{figure*}
	\begin{subfloat}[2 features]{
			\includegraphics[width=0.3\textwidth]{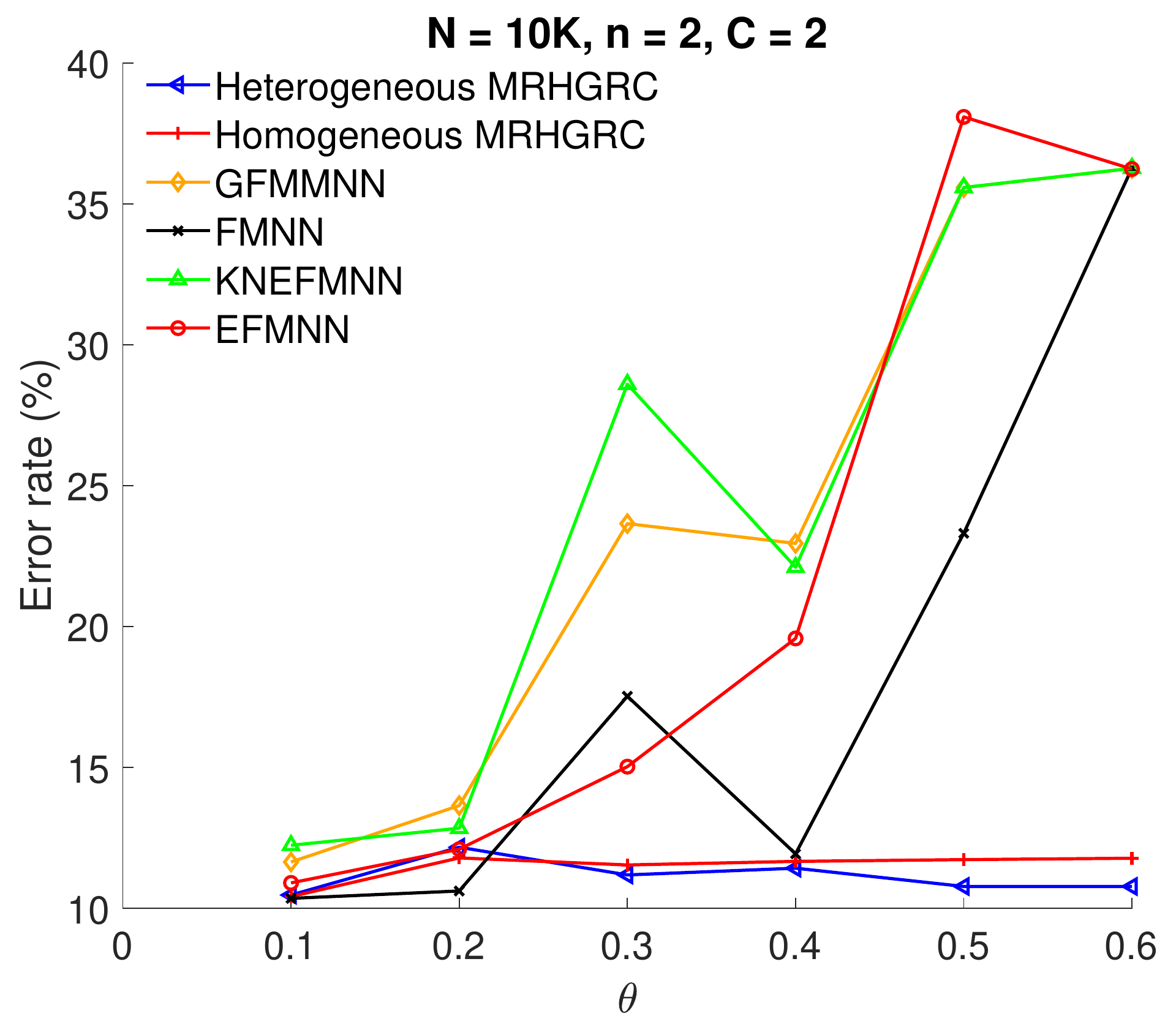}}
	\end{subfloat}
	\hfill%
	\begin{subfloat}[8 features]{
			\includegraphics[width=0.3\textwidth]{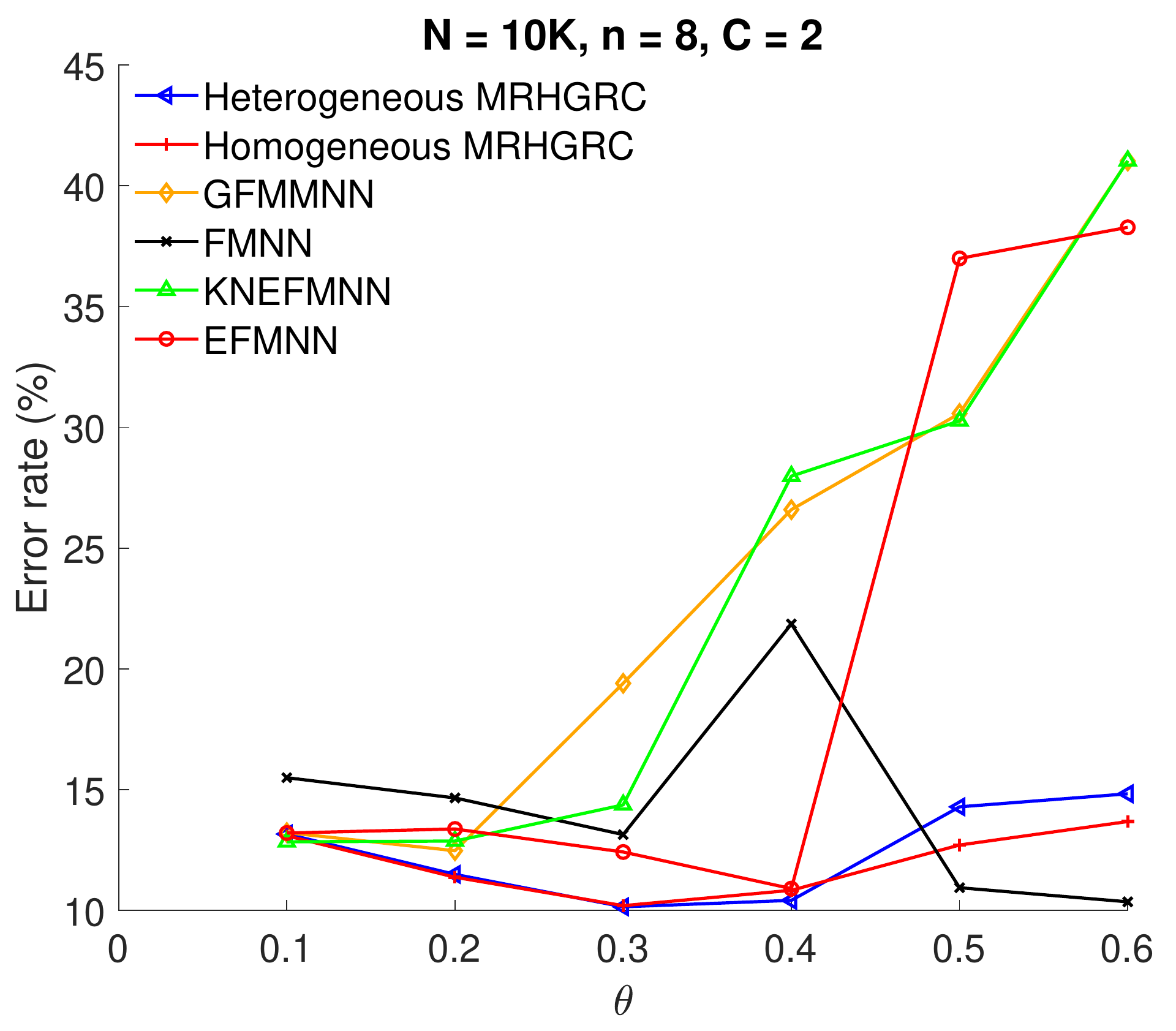}}
	\end{subfloat}
	\hfill%
	\begin{subfloat}[32 features]{
			\includegraphics[width=0.3\textwidth]{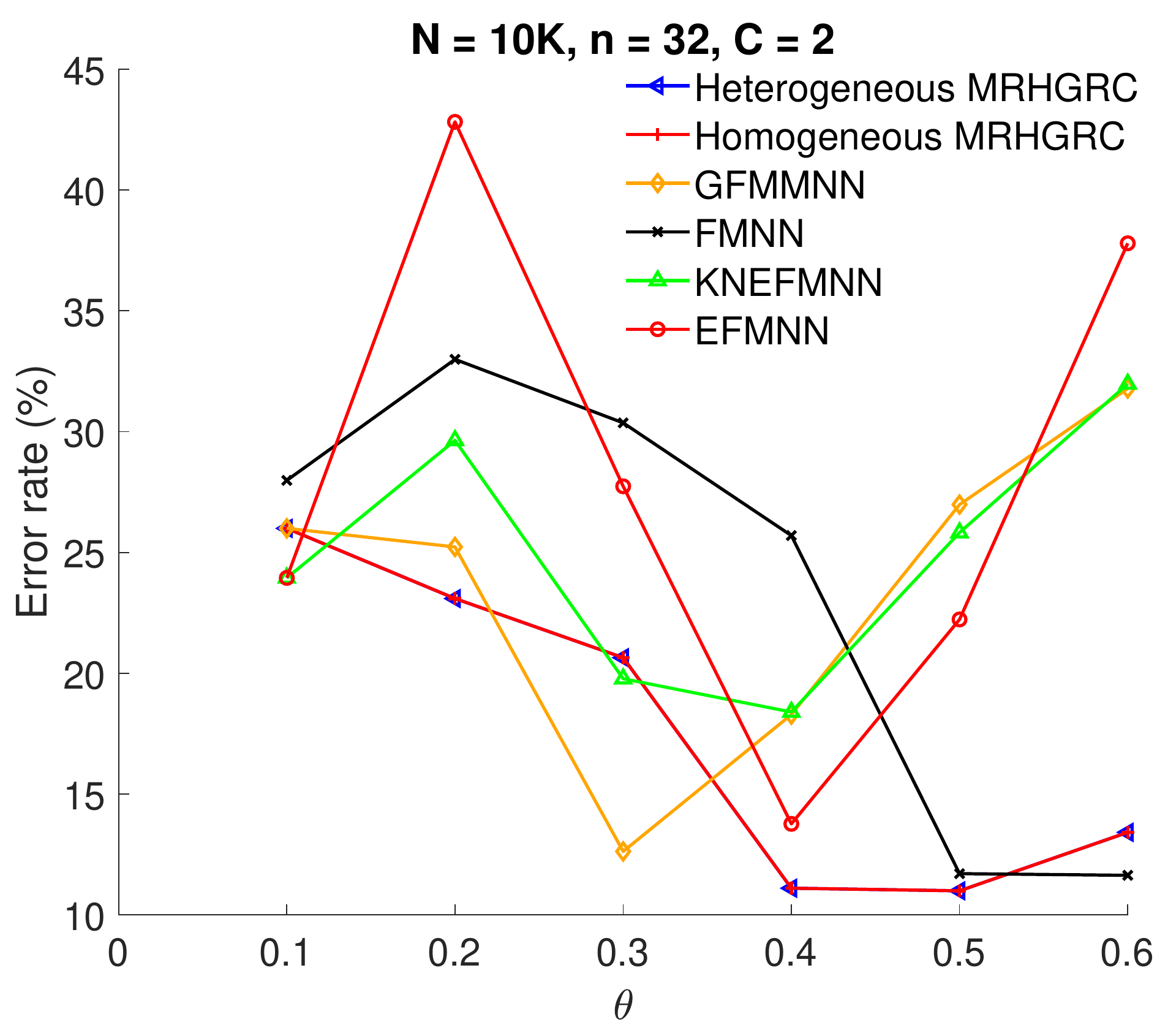}}
	\end{subfloat}
	\caption{The error rate of classifiers on synthetic linear boundary datasets with the different number of features.}
	\label{Fig.5}
\end{figure*}

\begin{table}
	\centering
	\caption{The Lowest Error Rates and Training Time of Classifiers on Synthetic Linear Boundary Datasets With Different Classes ($ N = 10K, n = 2$)}
	\label{Table.2}
	{\scriptsize
		\begin{tabular}{|l|l|r|r|c|c|r|}
			\hline
			\bfseries C          & \bfseries Algorithm            & \bfseries $\min E_{V}$ & \bfseries $\min E_{T}$ & \bfseries $\theta_{V}$ & \bfseries $\theta_{T}$ & \bfseries Time (s) \\ 
			\hline \hline
			\multirow{9}{*}{2}  & He-MRHGRC            & 10.25          & 10.467          & 0.1            & 0.1             & 1.1378              \\ \cline{2-7} 
			& Ho-MRHGRC            & 10.10          & 10.413          & 0.1            & 0.1             & 1.3215              \\ \cline{2-7} 
			& GFMM                 & 11.54          & 11.639          & 0.1            & 0.1             & 8.6718              \\ \cline{2-7} 
			& FMNN                 & 10.05          & 10.349          & 0.1            & 0.1             & 46.4789             \\ \cline{2-7} 
			& KNEFMNN              & 12.07          & 12.232          & 0.1            & 0.1             & 9.4459              \\ \cline{2-7} 
			& EFMNN                & 10.44          & 10.897          & 0.1            & 0.1             & 48.9892             \\ \cline{2-7} 
			& GNB                  & \textbf{9.85}           & \textbf{9.964}           & -              & -               & 0.5218              \\ \cline{2-7} 
			& SVM                  & 9.91           & 9.983           & -              & -               & 1.5468              \\ \cline{2-7} 
			& DT                   & 15.33          & 14.861          & -              & -               & 0.5405              \\ \hline
			\multirow{9}{*}{4}  & He-MRHGRC            & 19.76          & 19.884          & 0.4            & 0.4             & 1.0754              \\ \cline{2-7} 
			& Ho-MRHGRC            & 19.97          & 21.135          & 0.1            & 0.1             & 1.5231              \\ \cline{2-7} 
			& GFMM                 & 22.34          & 22.515          & 0.1            & 0.1             & 10.8844             \\ \cline{2-7} 
			& FMNN                 & 20.00          & 20.350          & 0.1            & 0.1             & 65.7884             \\ \cline{2-7} 
			& KNEFMNN              & 20.54          & 20.258          & 0.1            & 0.1             & 12.5618             \\ \cline{2-7} 
			& EFMNN                & 21.75          & 21.736          & 0.1            & 0.1             & 55.1921             \\ \cline{2-7} 
			& GNB                  & 19.35          & \textbf{19.075}          & -              & -               & 0.5492              \\ \cline{2-7} 
			& SVM                  & \textbf{19.34}          & 19.082          & -              & -               & 1.6912              \\ \cline{2-7} 
			& DT                   & 26.94          & 27.014          & -              & -               & 0.5703              \\ \hline
			\multirow{9}{*}{16} & He-MRHGRC            & 30.11          & 30.996          & 0.1            & 0.4             & 1.2686              \\ \cline{2-7} 
			& Ho-MRHGRC            & 28.70          & 30.564          & 0.1            & 0.1             & 1.8852              \\ \cline{2-7} 
			& GFMM                 & 32.66          & 33.415          & 0.1            & 0.1             & 18.0554             \\ \cline{2-7} 
			& FMNN                 & 29.78          & 31.035          & 0.1            & 0.1             & 69.6761             \\ \cline{2-7} 
			& KNEFMNN              & 33.42          & 34.670          & 0.1            & 0.1             & 22.3418             \\ \cline{2-7} 
			& EFMNN                & 31.80          & 33.239          & 0.1            & 0.1             & 76.0920             \\ \cline{2-7} 
			& GNB                  & \textbf{27.12}          & 28.190          & -              & -               & 0.5764              \\ \cline{2-7} 
			& SVM                  & 27.29          & \textbf{28.103}          & -              & -               & 1.6455              \\ \cline{2-7} 
			& DT                   & 38.813         & 39.644          & -              & -               & 0.6023              \\ \hline
		\end{tabular}
	}
\end{table}

Fig. \ref{Fig.4} shows the change in error rates of fuzzy min-max classifiers with a different number of classes on the testing sets. It can be easily seen that the error rates of our method are lowest compared to other fuzzy min-max neural networks on all multi-class synthetic datasets at high abstraction levels of granular representations. At high abstraction levels, the error rates of other fuzzy min-max neural networks increase rapidly, while the error rate of our classifier still maintains the stability. In addition, the error rates of our method also slowly augment in contrast to the behaviors of other considered types of fuzzy min-max neural networks when increasing the abstraction level of granular representations. These facts demonstrate the efficiency of our proposed method on multi-class datasets. The lowest error rates of classifiers on validation and testing sets, as well as total training time, are shown in Table \ref{Table.2}. It is observed that the predictive accuracy of our method outperforms all considered types of fuzzy min-max classifiers and decision tree, but it cannot overcome the Gaussian Naive Bayes and support vector machine methods. The training time of our method is faster than other fuzzy min-max neural networks and support vector machines on the considered multi-class synthetic datasets.

\textbf{\textit{Increase the number of features:}}

To generate the multi-dimensional synthetic datasets with the number of samples $ N = 10K $ and the number of classes $ C = 2$, we used the similar settings as in generation of datasets with different number of samples. The means of classes are $ \mu_1 = [0, \ldots, 0]^T, \mu_2 = [2.56, 0, \ldots, 0]^T $, and the covariance matrices are as follows:

\begin{equation*}
\scriptsize
\Sigma_1 = \Sigma_2 = \left[ \begin{array}{ccc}
1& \ldots & 0\\
\vdots &\ddots& \vdots\\
0& \ldots &1\\
\end{array}
\right]
\end{equation*}
The size of each expression corresponds to the number of dimensions $n$ of the problem. Fukunaga \cite{Fukunaga90} stated that the general Bayes error of 10\% and this Bayes error stays the same even when $n$ changes.

\begin{table}
	\centering
	\caption{The Lowest Error Rates and Training Time of Classifiers on Synthetic Linear Boundary Datasets With Different Features ($ N = 10K, C = 2$)}
	\label{Table.3}
	{\scriptsize
		\begin{tabular}{|l|l|r|r|c|c|r|}
			\hline
			\bfseries n          & \bfseries Algorithm            & \bfseries $\min E_{V}$ & \bfseries $\min E_{T}$ & \bfseries $\theta_{V}$ & \bfseries $\theta_{T}$ & \bfseries Time (s) \\ 
			\hline \hline
			\multirow{9}{*}{2}  & He-MRHGRC             & 10.250         & 10.467         & 0.1            & 0.1             & 1.1378              \\ \cline{2-7} 
			& Ho-MRHGRC             & 10.100         & 10.413         & 0.1            & 0.1             & 1.3215              \\ \cline{2-7} 
			& GFMM                 & 11.540         & 11.639          & 0.1            & 0.1             & 8.6718              \\ \cline{2-7} 
			& FMNN                 & 10.050         & 10.349          & 0.1            & 0.1             & 46.4789             \\ \cline{2-7} 
			& KNEFMNN              & 12.070         & 12.232          & 0.1            & 0.1             & 9.4459              \\ \cline{2-7} 
			& EFMNN                & 10.440         & 10.897          & 0.1            & 0.1             & 48.9892             \\ \cline{2-7} 
			& GNB                  & \textbf{9.850}          & \textbf{9.964}           & -              & -               & 0.5218              \\ \cline{2-7} 
			& SVM                  & 9.910          & 9.983           & -              & -               & 1.5468              \\ \cline{2-7} 
			& DT                   & 15.330         & 14.861          & -              & -               & 0.5405              \\ \hline
			\multirow{9}{*}{8}  & He-MRHGRC            & 10.330         & 10.153          & 0.3            & 0.3             & 21.9131             \\ \cline{2-7} 
			& Ho-MRHGRC            & 10.460         & 10.201          & 0.3            & 0.3             & 23.0554             \\ \cline{2-7} 
			& GFMM                 & 12.170         & 12.474          & 0.1            & 0.2             & 196.0682            \\ \cline{2-7} 
			& FMNN                 & 10.250         & 10.360          & 0.6            & 0.6             & 302.8683            \\ \cline{2-7} 
			& KNEFMNN              & 12.720         & 12.844          & 0.1            & 0.1             & 618.2524            \\ \cline{2-7} 
			& EFMNN                & 11.300         & 10.907          & 0.4            & 0.4             & 579.3113            \\ \cline{2-7} 
			& GNB                  & \textbf{9.940}          & \textbf{9.919}           & -              & -               & 0.5915              \\ \cline{2-7} 
			& SVM                  & 9.980          & 9.927           & -              & -               & 2.0801              \\ \cline{2-7} 
			& DT                   & 15.383         & 15.087          & -              & -               & 0.6769              \\ \hline
			\multirow{9}{*}{32} & He-MRHGRC            & 11.070         & 10.995          & 0.5            & 0.5             & 226.3193            \\ \cline{2-7} 
			& Ho-MRHGRC            & 11.070         & 10.995          & 0.5            & 0.5             & 226.0611            \\ \cline{2-7} 
			& GFMM                 & 12.390         & 12.625          & 0.3            & 0.3             & 847.6977            \\ \cline{2-7} 
			& FMNN                 & 11.830         & 11.637          & 0.5            & 0.6             & 1113.6836           \\ \cline{2-7} 
			& KNEFMNN              & 17.410         & 18.395          & 0.1            & 0.4             & 837.9571            \\ \cline{2-7} 
			& EFMNN                & 13.890         & 13.766          & 0.4            & 0.4             & 1114.4976           \\ \cline{2-7} 
			& GNB                  & 10.280         & 10.088          & -              & -               & 0.7154              \\ \cline{2-7} 
			& SVM                  & \textbf{10.220}         & \textbf{10.079}          & -              & -               & 4.5937              \\ \cline{2-7} 
			& DT                   & 15.400         & 15.201          & -              & -               & 1.0960              \\ \hline
		\end{tabular}
	}
\end{table}

Fig. \ref{Fig.5} shows the change in the error rates with different levels of granularity on multi-dimensional synthetic datasets. In general, with a low number of dimensions, our method outperforms other fuzzy min-max neural networks. With high dimensionality and a small number of samples, the high levels of granularity result in high error rates, and misclassification results considerably drops when the value of $ \theta $ increases. The same trend also happens to the FMNN when its accuracy at $ \theta = 0.5 $ or $ \theta = 0.6 $ is quite high. Apart from the FMNN on high dimensional datasets, our proposed method is better than three other fuzzy min-max classifiers at high abstraction levels. Table \ref{Table.3} reports the lowest error rates of classifiers on validation and testing multi-dimensional sets as well as the total training time through six abstraction levels of granular representations. The training time of our method is much faster than other types of fuzzy min-max neural networks. Generally, the performance of our proposed method overcomes the decision tree and other types of fuzzy min-max neural networks, but its predictive results cannot defeat the Gaussian Naive Bayes and support vector machines. It can be observed that the best performance on validation and testing sets obtains at the same abstraction level of granular representations on all considered multi-dimensional datasets. This fact indicates that we can use the validation set to choose the best classifier at a given abstraction level among constructed models through different granularity levels.

\subsubsection{Non-linear Boundary}~\\
To generate non-linear boundary datasets, we set up the Gaussian means of the first class: $ \mu_1 = [-2, 1.5]^T, \mu_2 = [1.5, 1]^T $ and the Gaussian means of the second class: $ \mu_3 = [-1.5, 3]^T, \mu_4 = [1.5, 2.5]^T $. The covariance matrices for the first class $ \Sigma_1, \Sigma_2 $ and for the second class $ \Sigma_3, \Sigma_4 $ were established as follows:

\begin{equation*}
\scriptsize
\begin{split}
&\Sigma_1 = \left[ \begin{array}{cc}
0.5 & 0.05\\
0.05 & 0.4\\
\end{array}
\right], \Sigma_2 = \left[ \begin{array}{cc}
0.5 & 0.05\\
0.05 & 0.3\\
\end{array}
\right], \\
&\Sigma_3 = \left[ \begin{array}{cc}
0.5 & 0\\
0 & 0.5\\
\end{array}
\right], \Sigma_4 = \left[ \begin{array}{cc}
0.5 & 0.05\\
0.05 & 0.2\\
\end{array}
\right], 
\end{split}
\end{equation*}
The number of samples for each class was equal, and the generated samples were normalized to the range of [0, 1]. We created only a testing set including 100,000 samples and a validation set with 10,000 patterns. Three different training sets containing 10K, 100K, and 5M samples were used to train classifiers. We aim to evaluate the predictive results of our method on the non-linear boundary dataset when changing the sizes of the training set.

\begin{figure*}
	\begin{subfloat}[10K samples]{
			\includegraphics[width=0.3\textwidth]{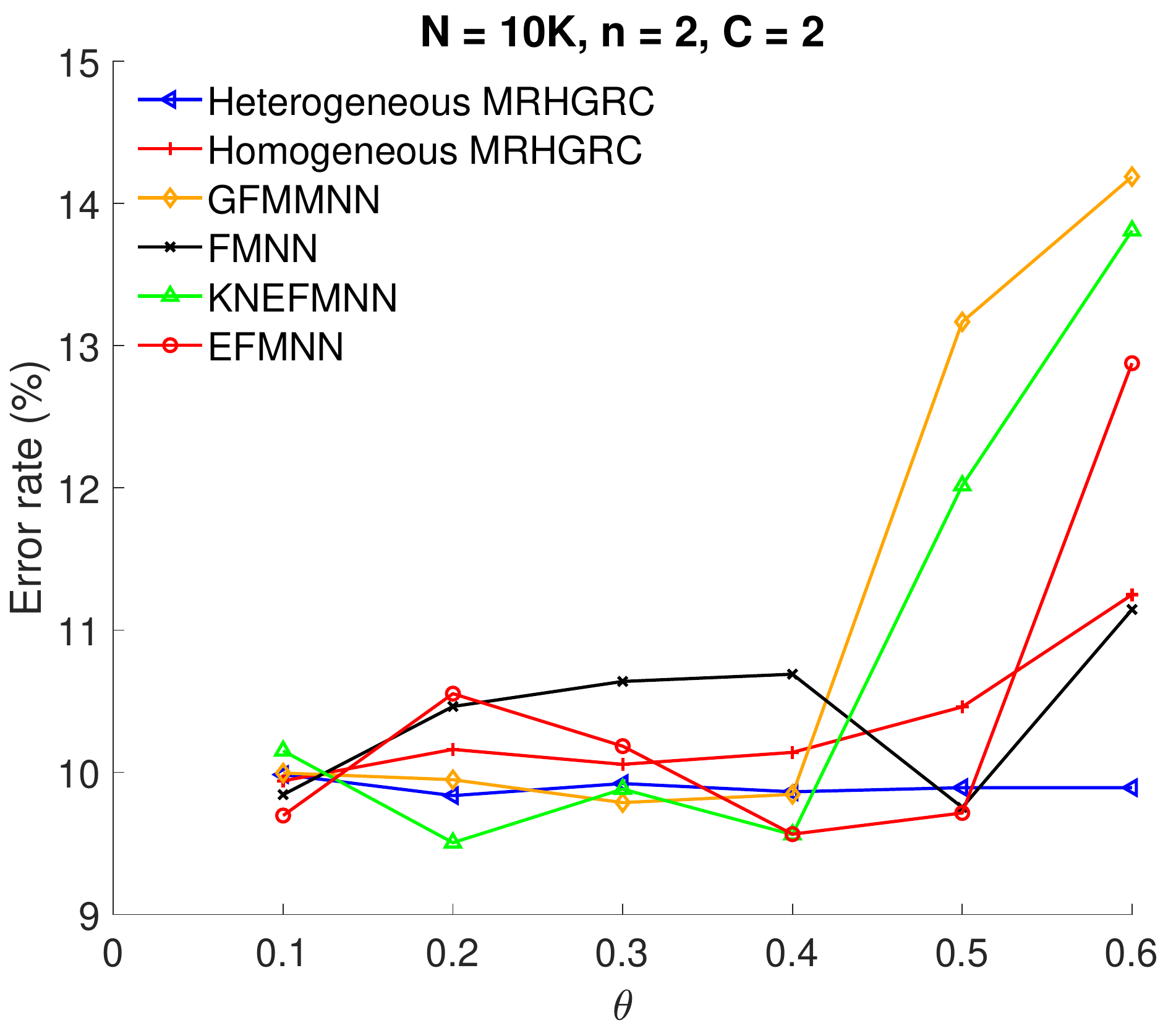}}
	\end{subfloat}
	\hfill%
	\begin{subfloat}[100K samples]{
			\includegraphics[width=0.3\textwidth]{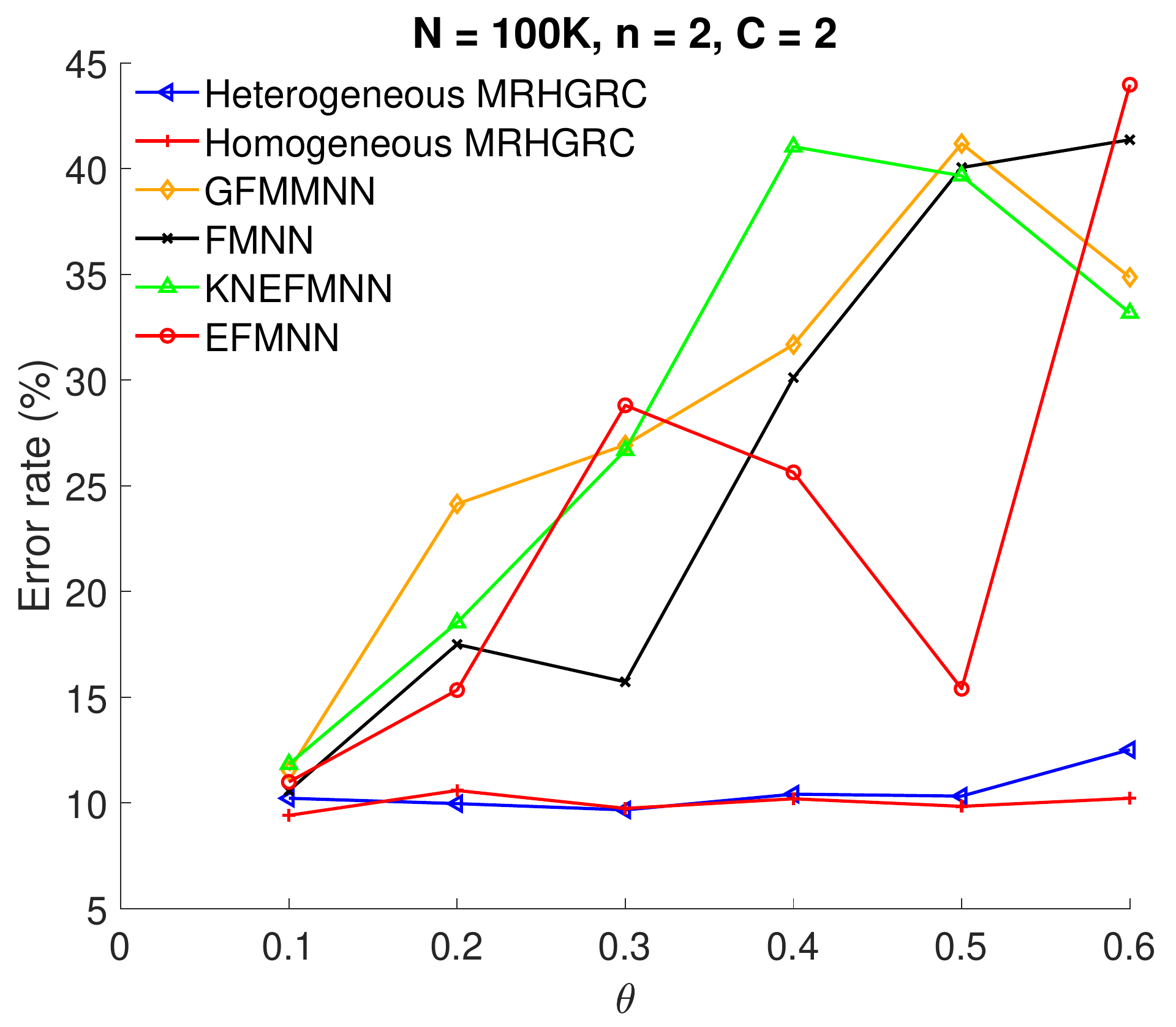}}
	\end{subfloat}
	\hfill%
	\begin{subfloat}[5M samples]{
			\includegraphics[width=0.3\textwidth]{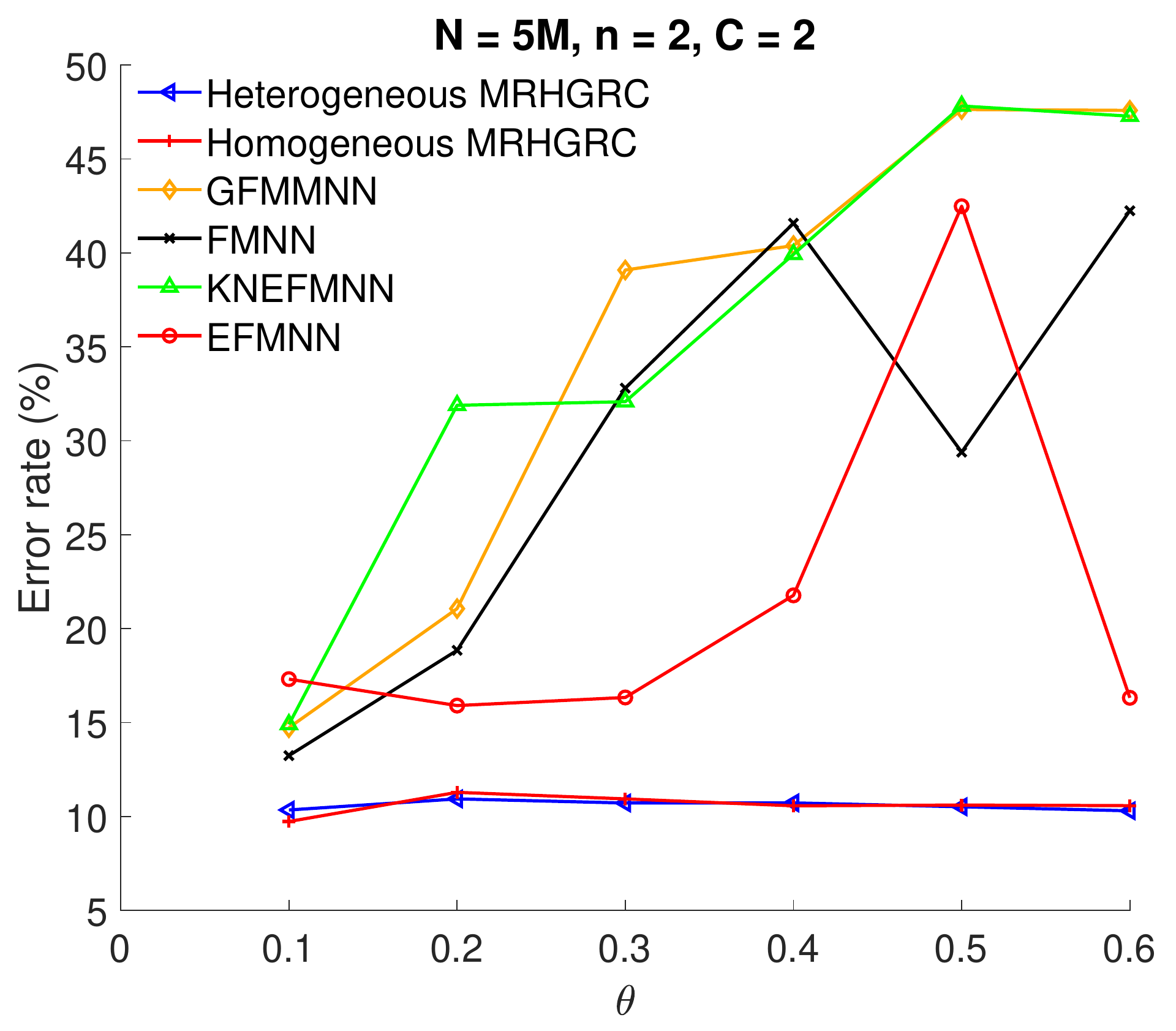}}
	\end{subfloat}
	\hfill%
	\caption{The error rate of classifiers on synthetic non-linear boundary datasets with the different number of samples.}
	\label{Fig.6}
\end{figure*}

Fig. \ref{Fig.6} shows the changes in the error rates through different levels of granularity of classifiers on non-linear boundary datasets. It can be observed that the error rates of our proposed method trained on the large-sized non-linear boundary datasets are better than those using other types of fuzzy min-max neural networks, especially at high abstraction levels of granular representations. While other fuzzy min-max neural networks show the increase in the error rates if the value of $ \theta $ grows up, our method is capable of maintaining the stability of predictive results even in the case of high abstraction levels. When the number of samples increases, the error rates of other fuzzy min-max classifiers usually rise, whereas the error rate in our approach only fluctuates a little. These results indicate that our approach may reduce the influence of overfitting because of constructing higher abstraction level of granular data representations using the learned knowledge from lower abstraction levels.

The best performance of our approach does not often happen at the smallest value of $ \theta $ on these non-linear datasets. Results regarding accuracy on validation and testing sets reported in Table \ref{Table.4} confirm this statement. These figures also illustrate the effectiveness of the processing steps in phase 2. Unlike the linear boundary datasets, our method overcomes the Gaussian Naive Bayes to become two best classifiers (along with SVM) among classifiers considered. Although SVM outperformed our approach, its runtime on large-sized datasets is much slower than our method. The training time of our algorithm is much faster than other types of fuzzy min-max neural networks and SVM, but it is still slower than Gaussian Naive Bayes and decision tree techniques.

\begin{table}
	\centering
	\caption{The Lowest Error Rates and Training Time of Classifiers on Synthetic Non-linear Boundary Datasets With Different Number of Samples ($ n = 2, C = 2 $)}
	\label{Table.4}
	{\scriptsize
		\begin{tabular}{|l|l|r|r|c|c|r|}
			\hline
			\bfseries N            & \bfseries Algorithm            & \bfseries $\min E_{V}$ & \bfseries $\min E_{T}$ & \bfseries $\theta_{V}$ & \bfseries $\theta_{T}$ & \bfseries Time (s) \\ 
			\hline \hline
			\multirow{9}{*}{10K}  & He-MRHGRC             & 9.950          & 9.836          & 0.2            & 0.2             & 0.9616              \\ \cline{2-7} 
			& Ho-MRHGRC             & 9.820          & 9.940          & 0.1            & 0.1             & 1.1070              \\ \cline{2-7} 
			& GFMM                 & 10.200         & 9.787           & 0.4            & 0.5             & 10.5495             \\ \cline{2-7} 
			& FMNN                 & 9.770          & 9.753           & 0.5            & 0.5             & 61.1130             \\ \cline{2-7} 
			& KNEFMNN              & 9.890          & 9.505           & 0.2            & 0.2             & 16.1099             \\ \cline{2-7} 
			& EFMNN                & \textbf{9.750}          & 9.565           & 0.1            & 0.4             & 60.6073             \\ \cline{2-7} 
			& GNB                  & 10.740         & 10.626          & -              & -               & 0.5218              \\ \cline{2-7} 
			& SVM                  & \textbf{9.750}          & \textbf{9.490}           & -              & -               & 1.5565              \\ \cline{2-7} 
			& DT                   & 14.107         & 13.831          & -              & -               & 0.5388              \\ \hline
			\multirow{9}{*}{100K} & He-MRHGRC             & 10.130         & 9.670          & 0.3            & 0.3             & 2.5310              \\ \cline{2-7} 
			& Ho-MRHGRC             & 9.910          & 9.412          & 0.1            & 0.1             & 2.3560              \\ \cline{2-7} 
			& GFMM                 & 11.810         & 11.520          & 0.1            & 0.1             & 44.7778             \\ \cline{2-7} 
			& FMNN                 & 10.880         & 10.575          & 0.1            & 0.1             & 588.4412            \\ \cline{2-7} 
			& KNEFMNN              & 12.470         & 11.836          & 0.1            & 0.1             & 42.9151             \\ \cline{2-7} 
			& EFMNN                & 11.020         & 10.992          & 0.1            & 0.1             & 485.7613            \\ \cline{2-7} 
			& GNB                  & 10.830         & 10.702          & -              & -               & 0.9006              \\ \cline{2-7} 
			& SVM                  & \textbf{9.650}          & \textbf{9.338}           & -              & -               & 93.4474             \\ \cline{2-7} 
			& DT                   & 14.277         & 13.642          & -              & -               & 1.1767              \\ \hline
			\multirow{9}{*}{5M}   & He-MRHGRC            & 10.370         & 10.306          & 0.1            & 0.6             & 91.7894             \\ \cline{2-7} 
			& Ho-MRHGRC            & \textbf{9.940}          & \textbf{9.737}           & 0.1            & 0.1             & 69.5106             \\ \cline{2-7} 
			& GFMM                 & 15.260         & 14.730          & 0.1            & 0.1             & 1927.6191           \\ \cline{2-7} 
			& FMNN                 & 13.160         & 13.243          & 0.1            & 0.1             & 53274.4387          \\ \cline{2-7} 
			& KNEFMNN              & 15.040         & 14.905          & 0.1            & 0.1             & 1551.5220           \\ \cline{2-7} 
			& EFMNN                & 15.660         & 15.907          & 0.2            & 0.2             & 54487.6978          \\ \cline{2-7} 
			& GNB                  & 10.840         & 10.690          & -              & -               & 22.9849             \\ \cline{2-7} 
			& SVM                  & N/A            & N/A             & -              & -               & N/A                 \\ \cline{2-7} 
			& DT                   & 13.790         & 13.645          & -              & -               & 49.9919             \\ \hline
		\end{tabular}
	}
\end{table}

\subsection{Performance of the Proposed Method on Real Datasets}
\begin{table*}[!ht]
	\centering
	\caption{The Real Datasets and Their Statistics}
	\label{Table.5}
	\begin{tabular}{|l|c|c|c|c|c|c|}
		\hline
		\bfseries Dataset & \bfseries \#Dimensions & \bfseries \#Classes & \bfseries \#Training samples & \bfseries \#Validation samples & \bfseries \#Testing samples & \bfseries Source \\
		\hline \hline
		Poker Hand & 10 & 10 & 25,010 & 50,000 & 950000 & LIBSVM \\
		\hline
		SensIT Vehicle & 100 & 3 & 68,970 & 9,852 & 19,706 & LIBSVM \\
		\hline
		Skin\_NonSkin & 3 & 2 & 171,540 & 24,260 & 49,257 & LIBSVM\\
		\hline
		Covtype & 54 & 7 & 406,709 & 58,095 & 116,208 & LIBSVM \\
		\hline
		White wine quality & 11 & 7 & 2,449 & 1,224 & 1,225 & Kaggle \\
		\hline
		PhysioNet MIT-BIH Arrhythmia & 187 & 5 & 74,421 & 13,133 & 21,892 & Kaggle \\
		\hline
		MAGIC Gamma Telescope & 10 & 2 & 11,887 & 3,567 & 3,566 & UCI \\
		\hline
		Letter & 16 & 26 & 15,312 & 2,188 & 2,500 & UCI \\
		\hline
		Default of credit card clients & 23 & 2 & 18,750 & 5,625 & 5,625 & UCI \\
		\hline
		MoCap Hand Postures & 36 & 5 & 53,104 & 9,371 & 15,620 & UCI \\
		\hline
		MiniBooNE & 50 & 2 & 91,044 & 12,877 & 26,143 & UCI \\
		\hline
		SUSY & 18 & 2 & 4,400,000 & 100,000 & 500,000 & UCI \\
		\hline
	\end{tabular}
\end{table*}

Aiming to attain the fairest comparison, we used 12 datasets with diverse ranges of the number of sizes, dimensions, and classes. These datasets were taken from the LIBSVM \cite{Chang11}, Kaggle \cite{Kaggle19}, and UCI repositories \cite{Dua17} and their properties are described in Table \ref{Table.5}. For the \textit{SUSY} dataset, the last 500,000 patterns were used for the test set as shown in \cite{Baldi14}. From the results of synthetic datasets, we can see that the performance of the multi-resolution hierarchical granular representation based classifier using the heterogeneous data distribution technique is more stable than that utilizing the homogeneous distribution method. Therefore, the experiments in the rest of this paper were conducted for only the heterogeneous classifier.  

\begin{table}
	\caption{The Change in the Number of Generated Hyperboxes Through Different Levels of Granularity of the Proposed Method}
	\label{Table.6}
	{\scriptsize
		\begin{tabular}{|L{2cm}||r|r|r|r|r|r|}
			\hline
			\multirow{2}{2cm}{\bfseries Dataset} & \multicolumn{6}{c|}{\bfseries $\theta$} \\ \cline{2-7}
			& \bfseries 0.1 & \bfseries 0.2 & \bfseries 0.3 & \bfseries 0.4 & \bfseries 0.5 & \bfseries 0.6 \\ 
			\hline \hline
			Skin\_NonSkin & 1012 & 248 & 127 & 85 & 64 & 51 \\ \hline
			Poker Hand & 11563 & 11414 & 10905 & 3776 & 2939 & 2610 \\ \hline
			Covtype & 94026 & 13560 & 5224 & 2391 & 1330 & 846 \\ \hline
			SensIT Vehicle & 5526 & 2139 & 1048 & 667 & 523 & 457 \\ \hline
			PhysioNet MIT-BIH Arrhythmia & 60990 & 26420 & 15352 & 8689 & 5261 & 3241 \\ \hline
			White wine quality & 1531 & 676 & 599 & 559 & 544 & 526 \\ \hline
			Default of credit card clients & 2421 & 529 & 337 & 76 & 48 & 29 \\ \hline
			Letter & 9236 & 1677 & 952 & 646 & 595 & 556 \\ \hline
			MAGIC Gamma Telescope & 1439 & 691 & 471 & 384 & 335 & 308 \\ \hline
			MiniBooNE & 444 & 104 & 24 & 10 & 6 & 6 \\ \hline
			SUSY & - & - & 26187 & 25867 & 16754 & 13017 \\ \hline
		\end{tabular}
	}
\end{table}

Table \ref{Table.6} shows the number of generated hyperboxes for the He-MRHGRC on real datasets at different abstraction levels of granular representations. It can be seen that the number of hyperboxes at the value of $ \theta = 0.6 $ is significantly reduced in comparison to those at $ \theta = 0.1 $. However, the error rates of the classifiers on testing sets at $ \theta = 0.6 $ do not change so much compared to those at $ \theta = 0.1 $. This fact is illustrated in Fig. \ref{Fig.7} and figures in the supplemental file. From these figures, it is observed that at the high values of maximum hyperbox size such as $ \theta = 0.5 $ and $ \theta = 0.6 $, our classifier achieves the best performance compared to other considered types of fuzzy min-max neural networks. We can also observe that the prediction accuracy of our method is usually much better than that using other types of fuzzy min-max classifiers on most of the data granulation levels. The error rate of our classifier regularly increases slowly with the increase in the abstraction level of granules, even in some cases, the error rate declines at a high abstraction level of granular representations. The best performance of classifiers on validation and testing sets, as well as training time through six granularity levels, are reported in the supplemental file.

\begin{figure}
	\centering        
	\includegraphics[width=0.8\linewidth]{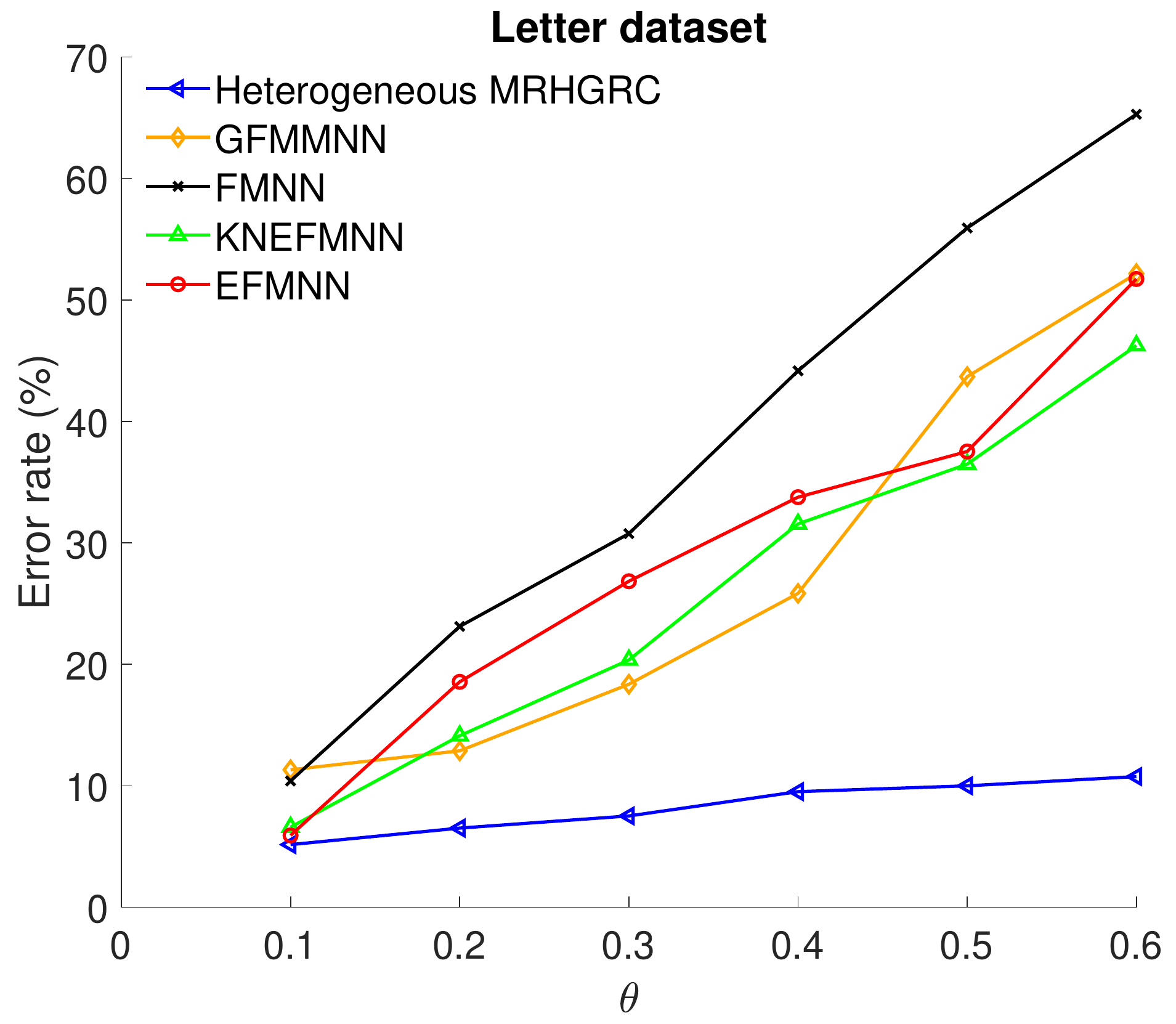}
	\caption{The error rate of classifiers on the \textit{Letter} datasets through data abstraction levels.}
	\label{Fig.7}
\end{figure}

Although our method cannot achieve the best classification accuracy on all considered datasets, its performance is located in the top 2 for all datasets. The Gaussian Naive Bayes classifiers obtained the best predictive results on synthetic linear boundary datasets, but it fell to the last position and became the worst classifier on real datasets because real datasets are highly non-linear. On datasets with highly non-linear decision boundaries such as \textit{covtype}, \textit{PhysioNet MIT-BIH Arrhythmia}, and \textit{MiniBooNE}, our proposed method still produces the good predictive accuracy.

The training process of our method is much faster than other types of fuzzy min-max neural networks on all considered datasets. Notably, on some large-sized complex datasets such as \textit{covtype} and \textit{SUSY}, the training time of other fuzzy min-max classifiers is costly, but their accuracy is worse than our method, which takes less training time. Our approach is frequently faster than SVM and can deal with datasets with millions of samples, while the SVM approach cannot perform.

On many datasets, the best predictive results on validation and testing sets were achieved at the same abstraction level of granular representations. In the case that the best model on the validation set has different abstraction level compared to the best model on the testing set, the error rate on the testing set if using the best classifier on the validation set is also near the minimum error. These figures show that our proposed method is stable, and it can achieve a high predictive accuracy on both synthetic and real datasets.

\subsection{The Vital Role of the Pruning Process and the Use of Sample Centroids}
\begin{table*}[!ht]
	\centering
	\caption{The Role of the Pruning Process and the Use of Sample Centroids}
	\label{roleofpruning}
	{\scriptsize
		\begin{tabular}{|l|R{1.7cm}|R{1.5cm}|R{1.5cm}|R{1.5cm}|R{1.3cm}|R{1.3cm}|R{1.3cm}|R{1.3cm}|}
			\hline
			\multirow{2}{*}{\bfseries Dataset} & \multicolumn{2}{c|}{\bfseries Num hyperboxes} & \bfseries Error rate before pruning (\%) & \bfseries Error rate after pruning (\%) & \multicolumn{2}{R{2.6cm}|}{\bfseries No. of predicted samples using centroids before pruning} & \multicolumn{2}{R{2.6cm}|}{\bfseries No. of predicted samples using centroids after pruning} \\ \cline{2-3} \cline{6-9} 
			& \bfseries Before pruning & \bfseries After pruning & & & \bfseries Total & \bfseries Wrong & \bfseries Total & \bfseries Wrong \\ \hline \hline
			Skin\_NonSkin & 1,358 & 1,012 & 0.1726 & 0.0974 & 1,509 & 73 & 594 & 30 \\ \hline
			Poker Hand & 24,991 & 11,563 & 53.5951 & 49.8128 & 600,804 & 322,962 & 725,314 & 362,196 \\ \hline
			SensIT Vehicle & 61,391 & 5,526	& 23.6730 & 20.9073 & 2	& 1	& 0 & 0 \\ \hline
			Default of credit card clients & 9,256 & 2,421 & 22.3822 & 19.7689 & 662 & 291 & 312 & 127 \\ \hline
			Covtype & 95971 & 94026 & 7.7335 & 7.5356 & 2700 & 975 & 2213 & 783 \\ \hline
			PhysioNet MIT-BIH Arrhythmia & 61,419 & 60,990 & 3.6589 & 3.5492 & 49 & 9 & 48 & 8 \\ \hline
			MiniBooNE & 1,079 & 444 & 16.4289 & 13.9043 & 14,947 & 3,404 & 11,205 & 2,575 \\ \hline
			SUSY & 55,096 & 26,187 & 30.8548 & 28.3456 & 410,094 & 145,709 & 370,570 & 124,850 \\ \hline
		\end{tabular}
	}
\end{table*}

This experiment aims to assess the important roles of the pruning process and the use of sample centroids on the performance of the proposed method. The experimental results related to these issues are presented in Table \ref{roleofpruning}. It is easily observed that the pruning step contributes to significantly reducing the number of generated hyperboxes, especially in \textit{SensIT Vehicle}, \textit{Default of credit card clients}, \textit{SUSY} datasets. When the poorly performing hyperboxes are removed, the accuracy of the model increases considerably. These figures indicate the critical role of the pruning process with regards to reducing the complexity of the model and enhancing the predictive performance.

We can also see that the use of sample centroids and Euclidean distance may predict accurately from 50\% to 95\% of the samples located in the overlapping regions between different classes. The predictive accuracy depends on the distribution and complexity of underlying data. With the use of sample centroids, we do not need to use the overlap test and contraction process in phase 1 at the highest level of granularity. This strategy leads to accelerating the training process of the proposed method compared to other types of fuzzy min-max neural networks, especially in large-sized datasets such as \textit{covtype} or \textit{SUSY}. These facts point to the effectiveness of the pruning process and the usage of sample centroids on improving the performance of our approach in terms of both accuracy and training time.

\subsection{Ability to Handling Missing Values}
This experiment was conducted on two datasets containing many missing values, i.e., \textit{PhysioNet MIT-BIH Arrhythmia} and \textit{MoCap Hand Postures} datasets. The aim of this experiment is to demonstrate the ability to handle missing values of our method to preserve the uncertainty of input data without doing any pre-processing steps. We also generated three other training datasets from the original data by replacing missing values with the zero, mean, or median value of each feature. Then, these values were used to fill in the missing values of corresponding features in the testing and validation sets. The obtained results are presented in Table \ref{missingvalues}. The predictive accuracy of the classifier trained on the datasets with missing values cannot be superior to ones trained on the datasets imputed by the median, mean or zero values. However, the training time is reduced, and the characteristic of the proposed method is still preserved, in which the accuracy of the classifier is maintained at high levels of abstraction, and its behavior is nearly the same on both validation and testing sets. The replacement of missing values by other values is usually biased and inflexible in real-world applications. The capability of deducing directly from data with missing values ensures the maintenance of the online learning property of the fuzzy min-max neural network on the incomplete input data.

\begin{table}
	\caption{The Training Time and the Lowest Error Rates of Our Method on the Datasets With Missing Values}
	\label{missingvalues}
	{\scriptsize
		\begin{tabular}{|L{3.2cm}|R{1.3cm}|R{1.3cm}|R{1.3cm}|}
			\hline
			\bfseries Dataset & \bfseries Training time (s) & \bfseries $ \min E_{V} $ & \bfseries $ \min E_{T} $ \\ 
			\hline \hline
			Arrhythmia with replacing missing values by zero values   &  53,100.2895  & 3.0762 ($\theta = 0.1$) & 3.5492 ($\theta = 0.1$)   \\ \hline
			Arrhythmia with replacing missing values by mean values   &  60,980.5110  & 2.6879 ($\theta = 0.1$) & 3.3848 ($\theta = 0.1$) \\ \hline
			Arrhythmia with replacing missing values by median values &  60,570.4315  & 2.7031 ($\theta = 0.1$) & 3.2980 ($\theta = 0.2$)   \\ \hline
			Arrhythmia with missing values retained                   & 58,188.8138  & 2.6955 ($\theta = 0.1$) & 3.1473 ($\theta = 0.1$)   \\ \hline
			Postures with replacing missing values by zero values     & 5,845.9722   & 6.6482 ($\theta = 0.1$) & 7.7529 ($\theta = 0.4$)   \\ \hline
			Postures with replacing missing values by mean values     & 5,343.0038    & 8.5370 ($\theta = 0.1$) & 9.7631 ($\theta = 0.3$)   \\ \hline
			Postures with replacing missing values by median values   & 4,914.4475    & 8.4089 ($\theta = 0.1$) & 9.9936 ($\theta = 0.3$)    \\ \hline
			Postures with missing values retained                     & 2,153.8121    & 14.5662 ($\theta = 0.4$) & 13.7900 ($\theta = 0.4$)  \\ \hline
		\end{tabular}
	}
\end{table}

\subsection{Comparison to State-of-the-art Studies}
The purpose of this section is to compare our method with recent studies of classification algorithms on large-sized datasets in physics and medical diagnostics. The first experiment was performed on the \textit{SUSY} dataset to distinguish between a signal process producing super-symmetric particles and a background process. To attain this purpose, Baldi et al. \cite{Baldi14} compared the performance of a deep neural network with boosted decision trees using the area under the curve (AUC) metrics. In another study, Sakai et al. \cite{Sakai18} evaluated different methods of AUC optimization in combination with support vector machines to enhance the efficiency of the final predictive model. The AUC values of these studies along with our method are reported in Table \ref{Table.9}. It can be seen that our approach overcomes all approaches in Sakai's research, but it cannot outperform the deep learning methods and boosted trees on the considered dataset.

\begin{table}
	\centering
	\caption{The AUC Value of the Proposed Method and Other Methods on the \textit{SUSY} Dataset}
	\label{Table.9}
	{\scriptsize
		\begin{tabular}{|l|l|}
			\hline
			\bfseries Method & \bfseries AUC   \\ 
			\hline \hline
			Boosted decision tree \cite{Baldi14}                                       & 0.863 \\ \hline
			Deep neural network \cite{Baldi14}                                         & 0.876 \\ \hline
			Deep neural network with dropout  \cite{Baldi14}                           & \textbf{0.879} \\ \hline
			Positive-Negative and unlabeled data based AUC optimization \cite{Sakai18} & 0.647 \\ \hline
			Semi-supervised rankboost based AUC optimization \cite{Sakai18}            & 0.709 \\ \hline
			Semi-supervised AUC-optimized logistic sigmoid \cite{Sakai18}              & 0.556 \\ \hline
			Optimum AUC with a generative model \cite{Sakai18}                         & 0.577 \\ \hline
			He-MRHGRC (Our method)                                                      & 0.799 \\ \hline
		\end{tabular}
	}
\end{table}

\begin{table} [!ht]
	\centering
	\caption{The Accuracy of the Proposed Method and Other Methods on the \textit{PhysioNet MIT-BIH Arrhythmia} Dataset}
	\label{Table.10}
	{\scriptsize
		\begin{tabular}{|l|r|}
			\hline
			\bfseries Method & \bfseries Accuracy(\%) \\ \hline \hline
			Deep residual Convolutional neural network \cite{Kachuee18}       & 93.4     \\ \hline
			Augmentation + Deep convolutional neural network \cite{Acharya17} & 93.5     \\ \hline
			Discrete wavelet transform + SVM \cite{Martis13}                 & 93.8     \\ \hline
			Discrete wavelet transform + NN \cite{Martis13}                & 94.52    \\ \hline
			Discrete wavelet transform + Random Forest \cite{Li16}       & 94.6     \\ \hline
			Our method on the dataset with the missing values        & \textbf{96.85}    \\ \hline
			Our method on the dataset with zero padding              & 96.45    \\ \hline
		\end{tabular}
	}
\end{table}

The second experiment was conducted on a medical dataset (\textit{PhysioNet MIT-BIH Arrhythmia}) containing Electrocardiogram (ECG) signal used for the classification of heartbeats. There are many studies on ECG heartbeat classification such as deep residual convolution neural network \cite{Kachuee18}, a 9-layer deep convolutional neural network on the augmentation of the original data \cite{Acharya17}, combinations of a discrete wavelet transform with neural networks, SVM \cite{Martis13}, and random forest \cite{Li16}. The \textit{PhysioNet MIT-BIH Arrhythmia} dataset contains many missing values and above studies used the zero padding mechanism for these values. Our method can directly handle missing values without any imputations. The accuracy of our method on the datasets with missing values and zero paddings is shown in Table \ref{Table.10} along with results taken from other studies. It is observed that our approach on the dataset including missing values outperforms all other methods considered. From these comparisons, we can conclude that our proposed method is extremely competitive to other state-of-the-art studies published on real datasets.

\section{Conclusion and Future Work} \label{conclu}
This paper presented a method to construct classification models based on multi-resolution hierarchical granular representations using hyperbox fuzzy sets. Our approach can maintain good classification accuracy at high abstraction levels with a low number of hyperboxes. The best classifier on the validation set usually produces the best predictive results on unseen data as well. One of the interesting characteristics of our method is the capability of handling missing values without the need for missing values imputation. This property makes it flexible for real-world applications, where the data incompleteness usually occurs. In general, our method outperformed other typical types of fuzzy min-max neural networks using the contraction process for dealing with overlapping regions in terms of both accuracy and training time. Furthermore, our proposed technique can be scaled to large-sized datasets based on the parallel execution of the hyperbox building process at the highest level of granularity to form core hyperboxes from sample points rapidly. These hyperboxes are then refined at higher abstraction levels to reduce the complexity and maintain consistent predictive performance.

The patterns located in the overlapping regions are currently classified by using Euclidean distance to the sample centroids. Future work will focus on deploying the probability estimation measure to deal with these samples. The predictive results of the proposed method depend on the order of presentations of the training patterns because it is based on the online learning ability of the general fuzzy min-max neural network. In addition, the proposed method is sensitive to noise and outliers as well. In real-world applications, noisy data are frequently encountered; thus they can lead to serious stability issue. Therefore, outlier detection and noise removal are essential issues which need to be tackled in future work. Furthermore, we also intend to combine hyperboxes generated in different levels of granularity to build an optimal ensemble model for pattern recognition.

\section*{Acknowledgment}
T.T. Khuat acknowledges FEIT-UTS for awarding his PhD scholarships (IRS and FEIT scholarships).

\ifCLASSOPTIONcaptionsoff
  \newpage
\fi



\bibliographystyle{IEEEtran}
\bibliography{reference}
%

%
\vfill
\begin{IEEEbiography}[{\includegraphics[width=1.1in,height=1.3in,clip, keepaspectratio]{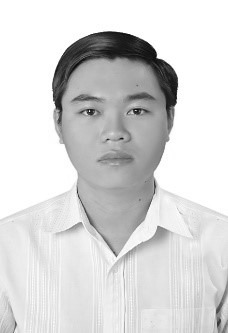}}] {Thanh Tung Khuat} received the B.E degree in Software Engineering from University of Science and Technology, Danang, Vietnam, in 2014. Currently, he is working towards the Ph.D. degree at the Advanced Analytics Institute, Faculty of Engineering and Information Technology, University of Technology Sydney, Ultimo, NSW, Australia. His research interests include machine learning, fuzzy systems, knowledge discovery, evolutionary computation, intelligent optimization techniques and applications in software engineering. He has authored and co-authored over 20 peer-reviewed publications in the areas of machine learning and computational intelligence.
\end{IEEEbiography}

\begin{IEEEbiography}[{\includegraphics[width=1in,height=1.25in,clip, keepaspectratio]{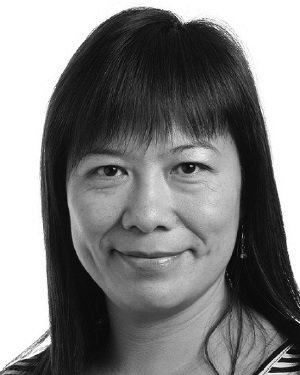}}]{Fang Chen} is the Executive Director Data Science and a Distinguished Professor with the University of Technology Sydney, Ultimo, NSW, Australia. She is a Thought Leader in AI and data science. She has created many world-class AI innovations while working with Beijing Jiaotong University, Intel, Motorola, NICTA, and CSIRO, and helped governments and industries utilising data and significantly increasing productivity, safety, and customer satisfaction. Through impactful successes, she gained many recognitions, such as the ITS Australia National Award 2014 and 2015, and NSW iAwards 2017. She is the NSW Water Professional of the Year 2016, the National and NSW Research, and the Innovation Award by Australian Water association. She was the recipient of the ``Brian Shackle Award" 2017 for the most outstanding contribution with international impact in the field of human interaction with computers and information technology. She is the recipient of the Oscar Prize in Australian science-Australian Museum Eureka Prize 2018 for Excellence in Data Science. She has 280 publications and 30 patents in 8 countries.
\end{IEEEbiography}

\begin{IEEEbiography}[{\includegraphics[width=1in,height=1.25in,clip]{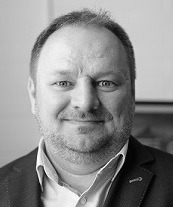}}]{Bogdan Gabrys} received the M.Sc. degree in electronics and telecommunication from Silesian Technical University, Gliwice, Poland, in 1994, and the Ph.D. degree in computer science from Nottingham Trent University, Nottingham, U.K., in 1998.
	
Over the last 25 years, he has been working at various universities and research and development departments of commercial institutions. He is currently a Professor of Data Science and a Director of the Advanced Analytics Institute at the University of Technology Sydney, Sydney, Australia. His research activities have concentrated on the areas of data science, complex adaptive systems, computational intelligence, machine learning, predictive analytics, and their diverse applications. He has published over 180 research papers, chaired conferences, workshops, and special sessions, and been on program committees of a large number of international conferences with the data science, computational intelligence, machine learning, and data mining themes. He is also a Senior Member of the Institute of Electrical and Electronics Engineers (IEEE), a Memebr of IEEE Computational Intelligence Society and a Fellow of the Higher Education Academy (HEA) in the UK. He is frequently invited to give keynote and plenary talks at international conferences and lectures at internationally leading research centres and commercial research labs. More details can be found at: http://bogdan-gabrys.com
\end{IEEEbiography}








\end{document}